\pdfoutput=1

\documentclass[11pt]{article}

\usepackage[]{emnlp2021}

\usepackage{times}
\usepackage{latexsym}

\usepackage[normalem]{ulem}
\usepackage{vcell}
\usepackage{amsmath}
\usepackage{amsfonts}
\usepackage{amssymb}
\usepackage{graphicx}
\usepackage{subfig}
\usepackage{float}
\usepackage{todonotes}

\usepackage[T1]{fontenc}

\usepackage[utf8]{inputenc}

\usepackage{microtype}

\title{TSDAE: Using Transformer-based Sequential Denoising Auto-Encoder for Unsupervised Sentence Embedding Learning}

\author{Kexin Wang, Nils Reimers, Iryna Gurevych \\
	Ubiquitous Knowledge Processing Lab (UKP-TUDA)\\
	Department of Computer Science, Technical University of Darmstadt\\
	\url{www.ukp.tu-darmstadt.de}}

\begin{document}
\maketitle
\begin{abstract}
Learning sentence embeddings often requires a large amount of labeled data. However, for most tasks and domains, labeled data is seldom available and creating it is expensive. In this work, we present a new state-of-the-art unsupervised method based on pre-trained Transformers and Sequential Denoising Auto-Encoder (TSDAE) which outperforms previous approaches by up to 6.4 points. It can achieve up to 93.1\% of the performance of in-domain supervised approaches. Further, we show that TSDAE is a strong domain adaptation and pre-training method for  sentence embeddings, significantly outperforming other approaches like Masked Language Model.\footnote{Code available at: \url{https://github.com/UKPLab/sentence-transformers/}}

A crucial shortcoming of previous studies is the narrow evaluation: Most work mainly evaluates on the single task of Semantic Textual Similarity (STS), which does not require any domain knowledge. It is unclear if these proposed methods generalize to other domains and tasks. We fill this gap and evaluate TSDAE and other recent approaches on four different datasets from heterogeneous domains.
\end{abstract}

\section{Introduction}
Sentence embedding techniques encode sentences into a fixed-sized, dense vector space such that semantically similar sentences are close. The most successful previous approaches like InferSent~\citep{conneau-EtAl:2017:EMNLP2017}, Universial Sentence Encoder (USE)~\citep{DBLP:conf/emnlp/CerYKHLJCGYTSK18} and SBERT~\citep{DBLP:conf/emnlp/ReimersG19} heavily relied on labeled data to train sentence embedding models. However, for most tasks and domains, labeled data is not available and data annotation is expensive. To overcome this limitation, unsupervised approaches have been proposed which learn to embed sentences just using an unlabeled corpus for training.

We propose a new approach: Transformer-based Sequential Denoising Auto-Encoder (TSDAE). It significantly outperforms previous methods via an encoder-decoder architecture. During training, TSDAE encodes corrupted sentences into fixed-sized vectors and requires the decoder to reconstruct the original sentences from this sentence embedding. For good reconstruction quality, the semantics must be captured well in the sentence embedding from the encoder. Later, at inference, we only use the encoder for creating sentence embeddings.

A crucial shortcoming of previous unsupervised approaches is the evaluation. Often, approaches are mainly evaluated on the Semantic Textual Similarity (STS) task from SemEval ~\citep{DBLP:conf/emnlp/LiZHWYL20,DBLP:journals/corr/abs-2006-03659,carlsson2021semantic,gao2021simcse}. As we argue in \autoref{sec:eval}, we perceive this as an insufficient evaluation. The STS datasets do not include sentences with domain specific knowledge, i.e., it remains unclear how methods will perform on more specific domains. Further, STS datasets have an artificial score distribution, and the performance on STS datasets does not correlate with downstream task performances \cite{DBLP:conf/coling/ReimersBG16}. In conclusion, it remains unclear, how well unsupervised sentence embedding methods will perform on domain specific tasks.

To answer this question, we compare TSDAE with previous unsupervised sentence embedding approaches on three different tasks (Information Retrieval, Re-Ranking and Paraphrase Identification), for heterogeneous domains and different text styles. We show that TSDAE can outperform other state-of-the-art unsupervised approaches by up to 6.4 points. TSDAE is able to perform on-par or even outperform existent supervised models like USE-large, which had been trained with a lot of labeled data from various datasets.

Further, we demonstrate that TSDAE works well for domain adaptation and as a pre-training task. We observe a significant performance improvement compared to other pre-training tasks like Masked Language Model (MLM).

Our contributions are three-fold:
\begin{itemize}
    \item We propose a novel unsupervised method, TSDAE based on denoising auto-encoders. We show that it outperforms the previous best approach by up to 6.4 points on diverse datasets.
    \item To the best of our knowledge, we are the first to compare recent unsupervised sentence embedding methods for various tasks on heterogeneous domains.
    \item TSDAE outperforms other methods including MLM by a large margin as a pre-training and domain adaptation method.
\end{itemize}

\section{Related Work}

\textbf{Supervised sentence embeddings} utilize labels for sentence pairs which provide the information about the relation between the sentences. Since sentence embeddings are usually applied to measure the similarity of a sentence pair, the most direct way is to label this similarity for supervised training~\citep{DBLP:journals/corr/HendersonASSLGK17}. Many studies also find that natural language inference (NLI), question answering and conversational context datasets can successfully be used to train sentence embeddings~\citep{conneau-EtAl:2017:EMNLP2017,DBLP:conf/emnlp/CerYKHLJCGYTSK18}. The recently proposed Sentence-BERT~\citep{DBLP:conf/emnlp/ReimersG19} introduced pre-trained Transformers to the field of sentence embeddings. Although high-quality sentence embeddings can be derived via supervised training, the labeling cost is a major obstacle for practical usage, especially for specialized domains.

\textbf{Unsupervised sentence embeddings} utilize only an unlabeled corpus during training. Recent work combined pre-trained Transformers with different training objectives to achieve state-of-the-art results on STS tasks. Among them,  Contrastive Tension (CT)~\citep{DBLP:journals/corr/abs-2006-03659} simply views the identical and different sentences as positive and negative examples, resp.\ and train two independent encoders; BERT-flow~\citep{DBLP:conf/emnlp/LiZHWYL20} trains model via debiasing embedding distribution towards Gaussian; SimCSE~\citep{gao2021simcse} is based on contrastive learning~\citep{DBLP:conf/cvpr/HadsellCL06,DBLP:conf/icml/ChenK0H20} and views the identical sentences with different dropout mask as the positive examples. For more details, please refer to \autoref{sec:exps}. All of them requires only independent sentences. By contrast, DeCLUTR~\citep{DBLP:journals/corr/abs-2006-03659} utilizes sentence-level contexts and requires long documents (2048 tokens at least) for training. This requirement is hardly met for many cases, e.g.\ tweets or dialogues. Thus, in this work we only consider methods which uses only single sentences during training.

Most previous work mainly evaluate only on Semantic Textual Similarity (STS) from the SemEval shared tasks. As we show in \autoref{sec:eval}, the unsupervised approaches perform much worse than the out-of-the-box supervised pre-trained models even though they were not specifically trained for STS. Further, a good performance on STS does not necessarily correlate with the performance on down-stream tasks~\citep{DBLP:conf/coling/ReimersBG16}. It remains unclear how these methods perform on specific tasks and domains. To answer this, we compare three recent powerful unsupervised methods based on pre-trained Transformers including CT, SimCSE, BERT-flow and our proposed TSDAE on different tasks of heterogeneous domains.

\section{Sequential Denoising Auto-Encoder} 
Although Sequential Denoising Auto-Encoder (SDAE)~\citep{DBLP:journals/jmlr/VincentLLBM10,Goodfellow-et-al-2016,DBLP:conf/naacl/HillCK16} is a popular unsupervised method in machine learning, how to combine it with pre-trained Transformers for learning sentence embeddings remains unclear. In this section, we first introduce the training objective of TSDAE and then give the optimal configuration of TSDAE.

\begin{figure}[t]
  \centering
  \includegraphics[width=40mm]{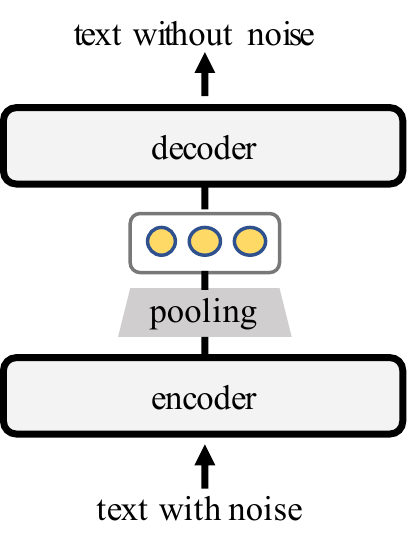}
  \caption{Architecture of TSDAE.}
  \label{fig:architecture}
\end{figure}

\subsection{Training Objective}
\autoref{fig:architecture} illustrates the architecture of TSDAE. TSDAE train sentence embeddings by adding a certain type of noise (e.g.\ deleting or swapping words) to input sentences, encoding the damaged sentences into fixed-sized vectors and then reconstructing the vectors into the original input. Formally, the training objective is:
\begin{align*}
J_{\mathrm{SDAE}}(\theta) &= \mathbb{E}_{x\sim D}[\log P_\theta(x|\tilde{x})]  \\
         &= \mathbb{E}_{x\sim D}[\sum_{t=1}^{l} \log P_\theta(x_t|\tilde{x})] \\
         &= \mathbb{E}_{x\sim D}[\sum_{t=1}^l \log \frac{\exp(h_t^Te_t)}{\sum_{i=1}^{N}\exp(h_t^Te_i)}]    
\end{align*}
where $D$ is the training corpus, $x=x_1x_2\cdots x_l$ is the input sentence with $l$ tokens, $\tilde{x}$ is the corresponding damaged sentence, $e_t$ is the word embedding of $x_t$, $N$ is the vocabulary size and $h_t$ is the hidden state at decoding step $t$. 

An important difference to original transformer encoder-decoder setup presented in~\citet{DBLP:conf/nips/VaswaniSPUJGKP17} is the information available to the decoder: Our decoder decodes only from a fixed-size sentence representation produced by the encoder. It does not have access to all contextualized word embeddings from the encoder. This modification introduces a bottleneck, that should force the encoder to produce a meaningful sentence representation.

\subsection{TSDAE}

The model architecture of TSDAE is a modified encoder-decoder Transformer where the key and value of the cross-attention are both confined to the sentence embedding only. Formally, the formulation of the modified cross-attention is:
\begin{align*}
&H^{(k)} = \mathrm{Attention}(H^{(k-1)}, [s^T], [s^T]) \\
&\mathrm{Attention}(Q, K, V) = \mathrm{softmax}\left(\frac{QK^T}{\sqrt{d}}\right)V
\end{align*}
where $H^{(k)}\in \mathbb{R}^{t\times d}$ is the decoder hidden states within $t$ decoding steps at the $k$-th layer, $d$ is the size of the sentence embedding, $[s^T]\in \mathbb{R}^{1\times d}$ is a one-row matrix including the sentence embedding vector and $Q$, $K$ and $V$ are the query, key and value, respectively. 
By exploring different configurations on the  STS benchmark dataset~\citep{cer-etal-2017-semeval}, we discover that the best combination is: (1) adopting deletion as the input noise and setting the deletion ratio to 0.6, (2) using the output of the \texttt{[CLS]} token as fixed-sized sentence representation (3) tying the encoder and decoder parameters during training. For the detailed tuning process, please refer to \autoref{sec:TSDAE_config}.

\section{Evaluation} \label{sec:eval}

Previous unsupervised sentence embedding learning approaches~\citep{DBLP:journals/corr/abs-2006-03659,carlsson2021semantic,DBLP:conf/emnlp/LiZHWYL20,su2021whitening, gao2021simcse} primarily evaluated on the task of Semantic Textual Similarity (STS) with data from SemEval using Pearson or Spearman's rank correlation. 

We find the (sole) evaluation on STS problematic. As shown in \cite{DBLP:conf/coling/ReimersBG16}, performance on the STS dataset does not correlate with downstream task performance, i.e.\ an approach working well on the STS tasks must not be a good choice for downstream tasks. We confirm this with our experiments, the performance on the STS tasks does not correlate with the performance on other (real-world) tasks. See \autoref{sec:STS_results} for more details on this.

This has multiple reasons: First, the STS datasets consists of sentences which do not require domain-specific knowledge, they are primarily from news and image captions. It is unclear how approaches will work for domain-specific tasks. Second, the STS datasets have an artificial score distribution - dissimilar and similar pairs appear roughly equally. For most real-word tasks, there is an extreme skew and only a tiny fraction of pairs are considered similar. Third, to perform well on the STS datasets, a method must rank dissimilar pairs and similar pairs equally well. In contrast, most real-world tasks, like duplicate questions detection, related paper finding, or paraphrase mining, only require to identify the few similar pairs out of a pool of millions of irrelevant combinations. 

A further shortcoming of previous evaluation setups is just testing the case of unsupervised learning, ignoring labeled data that potentially exists. In many scenarios, some labeled data exists, either directly from the specific task or from other (similar) tasks. A good approach should also work if some labeled data is available.

Hence, we propose to evaluate unsupervised sentence embedding approaches in following three setups:

\textbf{Unsupervised Learning:} We assume we just have unlabeled sentences from the target task and tune our approaches based on these sentences.

\textbf{Domain Adaptation:} We assume we have unlabeled sentences from the target task and labeled sentences from NLI \cite{DBLP:conf/emnlp/BowmanAPM15, DBLP:conf/naacl/WilliamsNB18} and STS benchmark \cite{cer-etal-2017-semeval} datasets. We test two setups: 1) Training on NLI+STS data, then unsupervised training to the target domain, 2) Unsupervised training on the target domain, then supervised training on NLI + STS.

\textbf{Pre-Training:} We assume we have a larger collection of unlabeled sentences from the target task and a smaller set of labeled sentences from the target task.


\begin{table*}[t]
\centering
\resizebox{14cm}{!}{
\begin{tabular}{|l|c|c|c|c|c|c|c|} 
\hline
\textbf{Dataset}             & \textbf{Task}          & \textbf{\#queries}     & \begin{tabular}[c]{@{}c@{}}\textbf{Avg.}\\\textbf{\#relevant} \end{tabular} & \begin{tabular}[c]{@{}c@{}}\textbf{Avg.}\\\textbf{\#candidates} \end{tabular} & \begin{tabular}[c]{@{}c@{}}\textbf{Avg.} \\\textbf{length} \end{tabular} & \begin{tabular}[c]{@{}c@{}}\textbf{Size of} \\\textbf{unsupervised}\\\textbf{training set} \end{tabular} & \begin{tabular}[c]{@{}c@{}}\textbf{Size of} \\\textbf{supervised} \\\textbf{training set} \end{tabular}  \\ 
\hline
\vcell{AskUbuntu}   & \vcell{RR}    & \vcell{200}   & \vcell{5.9/5.4}                                           & \vcell{20}                                                  & \vcell{9.2}                                            & \vcell{165K}                                                                  & \vcell{23K}                                                                   \\[-\rowheight]
\printcelltop       & \printcelltop & \printcelltop & \printcelltop                                             & \printcelltop                                               & \printcelltop                                          & \printcelltop                                                                 & \printcelltop                                                                 \\
\vcell{CQADupStack} & \vcell{IR}    & \vcell{3K}    & \vcell{1.1/1.1}                                           & \vcell{39K}                                                 & \vcell{8.6}                                            & \vcell{44K}                                                                   & \vcell{13K}                                                                   \\[-\rowheight]
\printcelltop       & \printcelltop & \printcelltop & \printcelltop                                             & \printcelltop                                               & \printcelltop                                          & \printcelltop                                                                 & \printcelltop                                                                 \\
\vcell{SciDocs}     & \vcell{RR}    & \vcell{4K}    & \vcell{5}                                                 & \vcell{30}                                                  & \vcell{12.5}                                           & \vcell{312K}                                                                  & \vcell{380K}                                                                  \\[-\rowheight]
\printcelltop       & \printcelltop & \printcelltop & \printcelltop                                             & \printcelltop                                               & \printcelltop                                          & \printcelltop                                                                 & \printcelltop                                                                 \\
\hline
\end{tabular}
}

\resizebox{14cm}{!}{
\begin{tabular}{|l|c|c|c|c|c|c|} 
\hline
\textbf{Dataset}             & \textbf{Task}          & \textbf{\#paraphrase}  & \textbf{\#non-paraphrase} & \begin{tabular}[c]{@{}c@{}}\textbf{Avg.} \\\textbf{length} \end{tabular} & \begin{tabular}[c]{@{}c@{}}\textbf{Size of} \\\textbf{unsupervised}\\\textbf{training set} \end{tabular} & \begin{tabular}[c]{@{}c@{}}\textbf{Size of} \\\textbf{supervised} \\\textbf{training set} \end{tabular}  \\ 
\hline
\vcell{TwitterPara} & \vcell{PI}    & \vcell{--/2K} & \vcell{--/9K}    & \vcell{{13.9}}            & \vcell{{53K}}                                     & \vcell{{23K}}                                     \\[-\rowheight]
\printcelltop       & \printcelltop & \printcelltop & \printcelltop    & \printcelltop                                          & \printcelltop                                                                 & \printcelltop                                                                 \\
\hline
\end{tabular}}

\caption{Dataset statistics. The slash symbol `/' separates the numbers for development and test. Multiple sub-datasets are included in CQADupStack, SciDocs and TwitterPara. CQADupStack has one sub-dataset for each of the 12 forums. The avg. \#relevant, avg.\ \#candidates and avg.\ length are all general statistics without distinguishing the sub-datasets.}

\label{tbl:data_statistics}
\end{table*}

\subsection{Datasets}

We evaluate these three settings on different tasks from heterogeneous (specialized) domains. The tasks include Re-Ranking (RR), Information Retrieval (IR) and Paraphrase Identification (PI). In detail, the datasets used are as follows\footnote{The dataset splits and the evaluation toolkit are available at: \url{https://github.com/UKPLab/useb}}:

\textbf{AskUbuntu} (RR task) is a collection of user posts from the technical forum AskUbuntu~\citep{DBLP:conf/naacl/LeiJBJTMM16}. Models are required to re-rank 20 candidate questions according to the similarity given an input post. The candidates are obtained via BM25 term-matching~\citep{DBLP:conf/trec/RobertsonWJHG94}. The evaluation metric is Mean Average Precision (MAP).

\textbf{CQADupStack} (IR task) is a question retrieval dataset of forum posts on various topics from StackExchange \citep{DBLP:conf/adcs/HoogeveenVB15}. In detail, it has 12 forums including Android, English, gaming, geographic information system, Mathematica, physics, programmers, statistics, Tex, Unix, Webmasters and WordPress. Models are required to retrieve duplicate questions from a large candidate pool. The metric is MAP@100. We train a single model for all forums.

\textbf{TwitterPara} (PI task) consists of two similar datasets: the Twitter Paraphrase Corpus (PIT-2015)~\citep{DBLP:conf/semeval/XuCD15} and the Twitter News URL Corpus (noted as TURL)~\citep{DBLP:conf/emnlp/LanQHX17}. The dataset consists of pairs of tweets together with a crowd-annotated score if the pair is a paraphrase. The evaluation metric is Average Precision (AP) over the gold confidence scores and the similarity scores from the models.

\textbf{SciDocs} (RR task) is a benchmark consisting of multiple tasks about scientific papers~\citep{DBLP:conf/acl/CohanFBDW20}. In our experiments, we use the tasks of \textit{Cite}: Given a paper title, identify the titles the paper is citing;  \textit{Co-Cite (CC)}, \textit{Co-Read (CR)}, and \textit{Co-View (CV)}, for which we must find papers that are frequently co-cited/-read/-viewed for a given paper title. For all these tasks, given one query paper title, models are required to identify up to 5 relevant papers titles from up to 30 candidates. The negative examples were selected randomly. The evaluation metric is MAP.

For evaluation, sentences are first encoded into fixed-sized vectors and cosine similarity is used for sentence similarity. Since we focus on embeddings for sentences, we just use the titles from the  AskUbuntu, CQADupStack and SciDocs datasets. For the datasets with sub-datasets or sub-tasks including CQADupStack, TwitterPara and SciDocs, the final score is derived by averaging the scores from each sub-dataset or sub-task.

For unsupervised training, we just use the sentences from the training split without any labels.  The statistics for each dataset are shown in \autoref{tbl:data_statistics}.


\section{Experiments}
\label{sec:exps}
In this section, we compare our proposed TSDAE with other unsupervised counterparts and  out-of-the-box supervised pre-trained models on the above mentioned tasks. For comparison, we include three recent state-of-the-art unsupervised approaches: CT, SimCSE, and BERT-flow. We use the proposed hyper-parameters from the respective paper. Without other specification, \textit{BERT-base-uncased}\footnote{Results for other checkpoints is reported in \autoref{sec:other_checkpoints}} is used as the base Transformer model. To eliminate the influence of randomness, we report the scores averaged over 5 random seeds. For other details, please refer to \autoref{sec:experiment_settings}.

\subsection{Baseline Methods}
\label{sec:baseline_methods}
We compare the approaches against avg.\ GloVe embeddings~\citep{pennington2014glove} and Sent2Vec~\citep{DBLP:conf/naacl/PagliardiniGJ18}. The former generates sentence embeddings by averaging word embeddings trained on a large corpus from the general domain; the latter is also a bag-of-words model but trained on the in-domain unlabeled corpus. The unsupervised baseline of \textit{BERT-base-uncased} with mean pooling is also in comparison. We further compare against existent pre-trained models: Universial Sentence Embedding (USE)~\citep{DBLP:conf/acl/YangCAGLCAYTSSK20}, which was trained on multiple supervised datasets including NLI and community question answering. From the Sentence-Transformers package, we use \textit{SBERT-base-nli-v2} and \textit{SBERT-base-nli-stsb-v2}: These models were trained on SNLI + MultiNLI data using the Multiple-Negative Ranking Loss (MNRL)~\citep{DBLP:journals/corr/HendersonASSLGK17} and the Mean Square Error (MSE) loss on the STS benchmark train set. Further we include BM25 using Elasticsearch for comparison.

To better understand the relative performance of these unsupervised methods, we also train SBERT models in an in-domain supervised manner and view their scores as the upper bound. For AskUbuntu, CQADupStack and SciDocs, where the relevant sentence pairs are labeled, the in-domain SBERT models are trained with MNRL. MNRL is a cross-entropy loss with in-batch negatives. For a batch of relevant sentences pairs $\{x^{(i)}, y^{(i)}\}_{i=1}^M$, MNRL views the labeled pairs as positive and the other in-batch combinations as negative. Formally, the training objective for each batch is:
\begin{align*}
J_{\mathrm{MNRL}}&(\theta) = \\ &\frac{1}{M}\sum_{i=1}^M\log\frac{\exp{\sigma(f_{\theta}(x^{(i)}), f_{\theta}(y^{(i)}))}}{\sum_{j=1}^M\exp{\sigma(f_{\theta}(x^{(i)}), f_{\theta}(y^{(j)}))}}
\end{align*}where $\sigma$ is a certain similarity function for vectors and $f_{\theta}$ is the sentence encoder that embeds sentences. For TwitterPara, whose relevant scores are labeled, the MSE loss is adopted to train the in-domain models.

\subsection{MLM} 
Masked-Language-Model (MLM) is a fill-in-the-blank task originally introduced by BERT: Words are masked from the input and the transformer network must predict the missing words. We use the original setup in~\citet{DBLP:conf/naacl/DevlinCLT19} except the number of training steps (100K), the batch size (8) and the learning rate (5e-5). To derive a sentence embedding, we perform mean-pooling of the output token embeddings.

\subsection{Contrastive Tension (CT)}
CT~\citep{carlsson2021semantic} finetunes pre-trained Transformers in a contrastive-learning fashion. For each sentence, it construct a binary cross-entropy loss by viewing the same sentence as the relevant and samples $K$ random sentences as the irrelevant.  To make the training process stable, for each sentence pair $(a, b)$, CT uses two independent encoders $f_{\theta_1}$ and $f_{\theta_2}$ from the same initial parameter point to encode the sentence $a$ and $b$, respectively. Formally, the learning objective is:
\begin{align*}
J_{\mathrm{CT}}&(\theta_1, \theta_2) = \mathbb{E}_{(a, b)\sim D}[y\log\sigma(f_{\theta_1}(a)^Tf_{\theta_2}(b)) \\ 
                        & +(1-y)\log(1-\sigma(f_{\theta_1}(a)^Tf_{\theta_2}(b))]
\end{align*}
where $y\in \{0, 1\}$ represents whether sentence $a$ is identical to sentence $b$ and $\sigma$ is the Logistic function. Despite its simplicity, CT achieves state-of-the-art unsupervised performance on the Semantic Textual Similarity (STS) datasets.

\subsection{SimCSE}
Similar to CT, SimCSE~\citep{gao2021simcse} also views the identical sentences as the positive examples. The main difference is that SimCSE samples different dropout masks for the same sentence to generate a embedding-level positive pair and uses in-batch negatives. Thus, this learning objective is equivalent to feeding each batch of sentences to the shared encoder twice and applying the MNRL-loss. 

\subsection{BERT-flow}
Instead of fine-tuning the parameters of the pre-trained Transformers, BERT-flow~\citep{DBLP:conf/emnlp/LiZHWYL20} aims at fully exploiting the semantic information encoded by these pre-trained models themselves via distribution debiasing. The paper of BERT-flow claims that the BERT word embeddings are highly relevant to the word frequency, which in turn influences the hidden states via the Masked Language Modeling (MLM) pre-training. This finally leads to biased sentence embeddings generated by the pooling over these hidden states. To solve this problem, BERT-flow inputs the biased sentence embedding into a trainable flow network $f_\phi$~\citep{DBLP:conf/nips/KingmaD18} for debiasing via fitting a standard Gaussian distribution, while keeping the parameters of the BERT model unchanged. Formally, the training objective is:
\begin{align}
    &J_{\mathrm{BERT}\text{-}\mathrm{flow}}(\phi) = \mathbb{E}_{x\sim D}[\log p_{\mathcal{U}}(u)] \label{eq:berflow1} \\
                          & = \mathbb{E}_u[\log(p_{\mathcal{Z}}(f_\phi^{-1}(u))|\det \frac{\partial f_\phi^{-1}(u)}{\partial u}|)] \label{eq:berflow2}
\end{align}
where $u$ is the biased embedding of sentence $x$ and $z=f_\phi^{-1}(u)$ is the debiased sentence embedding which follows a standard Gaussian distribution. Equation~\ref{eq:berflow2} is derived by applying the change-of-variables theorem to Equation~\ref{eq:berflow1}.

As BERT-flow does not update the parameters of the underlying Transformer network, we just reports scores for BERT-flow for \textit{unsupervised learning} and \textit{domain adaptation NLI+STS $\to$ target task}. It is not suitable for the other evaluation setups we used. We re-implemented BERT-flow under the Pytorch framework, which can reproduce the reported results in the original paper.\footnote{Code available at: \url{https://github.com/UKPLab/pytorch-bertflow}}

\section{Results}
\label{sec:results}

\begin{table*}[t]
\centering
\resizebox{15.5cm}{!}{
\begin{tabular}{|l|c|c|ccc|ccccc|c|} 
\hline
\textbf{Method}            & \textbf{AskU.}                              & \textbf{CQADup.}                & \multicolumn{3}{c|}{\textbf{TwitterP.}}                                              & \multicolumn{5}{c|}{\textbf{SciDocs}}                                                                                    & \textbf{Avg.}                   \\ 
\cline{4-11}
\textbf{Sub-task/-dataset} & \multicolumn{1}{l|}{}              & \multicolumn{1}{l|}{}  & \textbf{TURL}                   & \textbf{PIT}                    & \multicolumn{1}{l|}{\textbf{Avg.}} & \textbf{Cite}           & \textbf{CC}             & \textbf{CR}                             & \textbf{CV}             & \multicolumn{1}{l|}{\textbf{Avg.}} & \multicolumn{1}{l|}{}  \\ 
\hline
\multicolumn{12}{|l|}{ \textit{Unsupervised learning based on BERT-base} }                                                                                                                                                                                                                                         \\ 
\hline
TSDAE             & {59.4}$^\dagger$  & {14.5}$^\dagger$                   & {76.8}$^\dagger$                   & 69.2                   & 73.0                      & {71.4}$^\dagger$  & {73.9}$^\dagger$  & {75.0}$^\dagger$                  & {75.6}$^\dagger$  & {74.0}$^\dagger$             & {55.2}$^\dagger$          \\ 
MLM         & 54.3                               & 11.7                    & 71.9                   & 69.7 & 70.8                      & 71.2           & 75.8           & 75.1                           & 76.2           & 74.6                      & 52.9                   \\ 
CT                & {56.3}                       & {13.3}           & 74.6                   & {70.4}  & {72.5}              & {63.4}   & {67.1}   & {70.1}  & {69.7}   & {67.6}              & {52.4}           \\
SimCSE           & 55.9 & 12.4 & 74.5 & 62.5 & 68.5 &      62.5 & 65.1 & 67.7 & 67.6 & 65.7  & 50.6\\
BERT-flow         & 53.7                               & 9.2                    & 72.8                   & 65.7 & 69.3                      & 61.3           & 62.8           & 66.7                           & 67.1           & 64.5                      & 49.2                   \\

\hline
\multicolumn{12}{|l|}{ \textit{Domain adaptation: NLI+STS $\to$ target task} }  \\
\hline
TSDAE  & 58.7 & 13.6 & 75.8 & 66.2 & 71 & {69.9}$^\dagger$ & {73.8}$^\dagger$ & {75}$^\dagger$ & {75.7}$^\dagger$ & {73.6}$^\dagger$ & {54.2}$^\dagger$ \\
MLM & 54.4 & 9.7 & 69.8 & 68.1 & 69 & 67.1 & 71.8 & 72.6 & 72.9 & 71.1 & 51.1 \\
CT & 57.9 & 14.2 & 75.6 & 70.6 & 73.1 & 62.3 & 66.2 & 68.5 & 68.9 & 66.5 & 52.9 \\
SimCSE & 56.6 & 13.8 & 73.4 & 65.9 & 69.7 & 61.8 & 63.7 & 67.01 & 66.7 & 64.8 & 51.2 \\
BERT-flow & 58.2 & 13.9 & 76.5 & 67.4 & 72 & 62.2 & 64.8 & 68.1 & 68 & 65.8 & 52.5 \\

\hline
\multicolumn{12}{|l|}{ \textit{Domain adaptation: target task $\to$ NLI+STS} } \\
\hline
TSDAE & {59.4}$^\dagger$ & {14.4}$^\dagger$ & 75.8 & \textbf{73.1}$^\dagger$ & \textbf{74.5}$^\dagger$ & \textbf{75.6}$^\dagger$ & \textbf{78.6}$^\dagger$ & \textbf{78.1}$^\dagger$ & \textbf{78.2}$^\dagger$ & \textbf{77.6}$^\dagger$ & \textbf{56.5}$^\dagger$ \\
MLM & \textbf{60.6} & 14.3 & 75.0 & 68.6 & 71.8 & 74.7 & 78.2 & 77.0 & 77.6 & 76.9 & 55.9 \\
CT & 56.4 & 13.4 & 75.9 & 68.9 & 72.4 & 66.5 & 69.6 & 70.6 & 72.2 & 69.7 & 53.0 \\
SimCSE & 56.2 & 13.1 & 75.5 & 67.3 & 71.4 & 65.5 & 68.5 & 70.0 & 71.4 & 68.9 & 52.4 \\

\hline
\multicolumn{12}{|l|}{ \textit{Other previous unsupervised approaches} }                                                                                                                                                                                                                               \\ 
\hline
BM25              & {53.4}                       & 13.3                    & 71.9                   & 70.5                   & 71.2                      & 58.9           & 61.3           & 67.3                           & 66.9           & 63.6                      & 50.4                   \\
Avg. GloVe        & 51.0                               & {10.0}           & {70.1}           & {52.1}           & {61.1}              & 58.8           & 60.6           & 64.2                           & 65.4           & 62.2                      & {46.1}           \\
Sent2Vec          & 49.0                               & 3.2                    & 47.5                   & 39.9                   & 43.7                      & {61.6}   & {66.0}   & {66.1}                   & {66.7}   & {65.1}              & 40.2                   \\ 
BERT-base-uncased & 48.5 & 6.5 & 69.1 & 61.7 & 65.4 & 59.4 & 65.1 & 65.4 & 68.6 & 64.6 & 46.3 \\
\hline
\multicolumn{12}{|l|}{ \textit{Out-of-the-box supervised pre-trained models} }                                                                                                                                                                                                                         \\ 
\hline
SBERT-base-nli-v2 & 53.4 & 11.8 & 75.4 & 69.9 & 72.7 & 66.8 & 70.0 & 70.7 & 72.8 & 70.1 & 52.0 \\
SBERT-base-nli-stsb-v2 & 54.5 & 12.9 & 75.9 & 68.5 & 72.2 & 66.2 & 69.2 & 69.9 & 72.3 & 69.4 & 52.3\\
USE-large         & (\textit{59.3})                       & (\textit{\textbf{15.9}})  & \textbf{77.1}  & 69.8           & 73.5     & 67.1   & 69.5   & 71.4                   & 72.6   & {70.2}              & {54.7}           \\ 
\hline
\multicolumn{12}{|l|}{ \textit{In-domain supervised training (upper bound)} }                                                                                                                                                                                                                          \\ 
\hline
SBERT-supervised  & 63.8                               & 16.3                   & 81.6                   & 75.8                   & 78.7                      & 90.4           & 91.2           & 86.2                           & 83.6           & 87.9                      & 61.6                   \\
\hline
\end{tabular}}
\caption{Evaluation using average precision. Results are averaged over 5 random seeds. The best results excluding the upper bound are bold. USE-large was trained with in-domain training data for AskUbuntu and CQADupStack (scores in italic). Our proposed TSDAE significantly outperforms other unsupervised and supervised out-of-the-box approaches.$\dagger$ marks the cases where TSDAE outperforms both CT and SimCSE in all 5 runs.}
\label{tbl:main_results}
\end{table*}

\textbf{Unsupervised learning:} The results in \autoref{tbl:main_results} show that TSDAE can outperform the previous best approach (CT) by up-to 6.4 points (on SciDocs on average) and 2.6 points on average over all tasks. Surprisingly, a simple Masked-Language-Modeling (MLM) approach with mean pooling, which performs badly when evaluated on STS data, is the second best unsupervised approach, outperforming more recent approaches like CT, SimCSE, and BERT-flow on the selected tasks. TSDAE and MLM both removes words from the input, forcing the network to produce robust embeddings. In contrast, the input sentences for CT and SimCSE are not modified, resulting in less stable embeddings. Our experiments also show that out-of-the-box pre-trained models (SBERT-base-nli-stsb-v2 and USE-large) achieve strong results on our tasks without any domain-specific fine-tuning, outperforming recent proposed unsupervised learning approaches.

\textbf{Domain Adaptation:} For all unsupervised methods, we find that first training on the target domain, and then training with labeled NLI+STS achieves better results than the opposite direction.  For all methods, we observe a performance increase compared to only training on the target domain. On average, the performance improves by 1.3 points for TSDAE, 3.0 points for MLM, 0.6 points for CT, and 1.8 points for SimCSE. CT and SimCSE perform in this setting only slightly better than the out-of-the-box model SBERT-base-nli-stsb-v2.

\begin{figure}[t]
  \centering
  \subfloat[AskUbuntu]{\includegraphics[width=39mm]{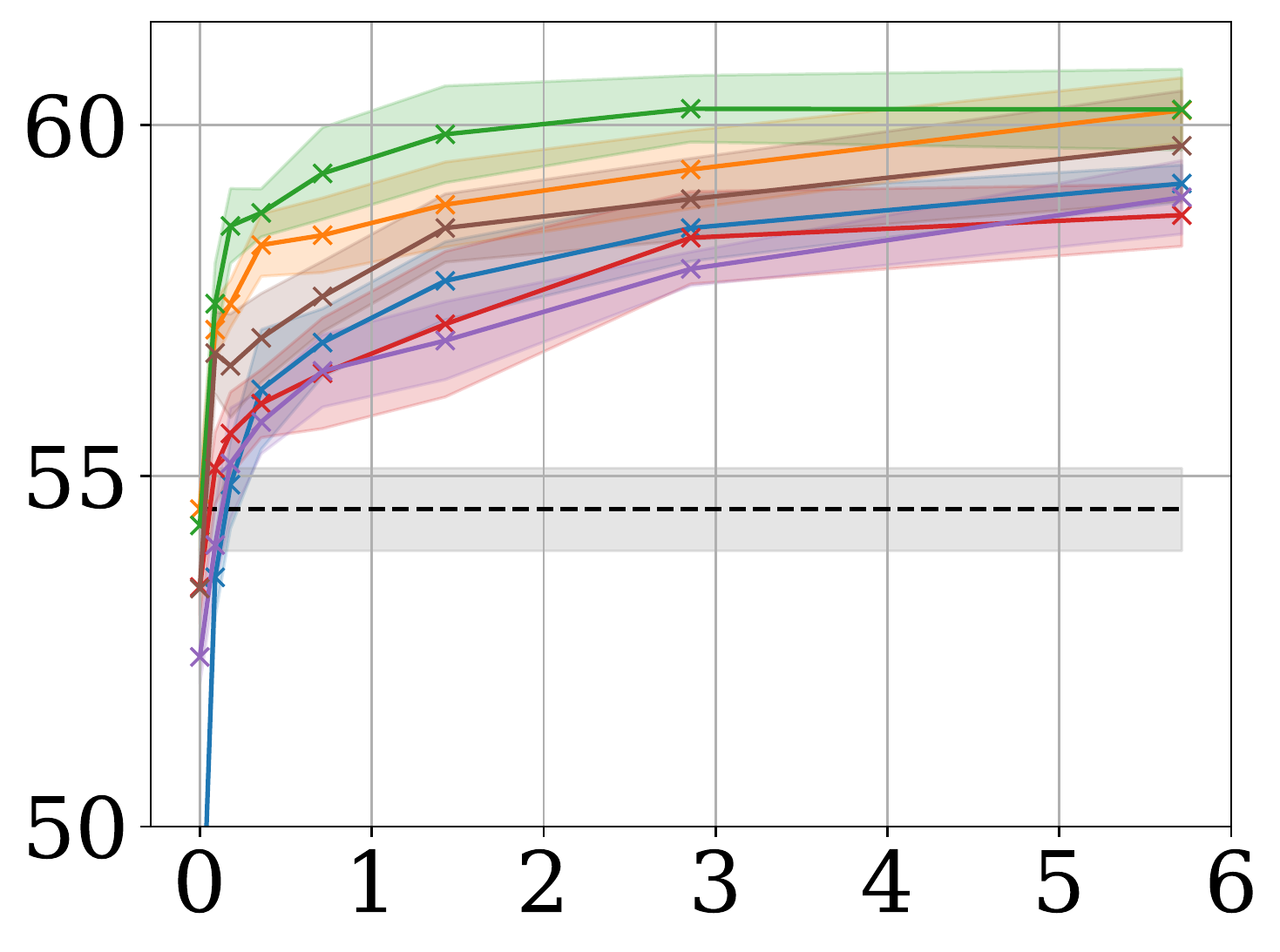}}
  \subfloat[CQADupStack]{\includegraphics[width=38mm]{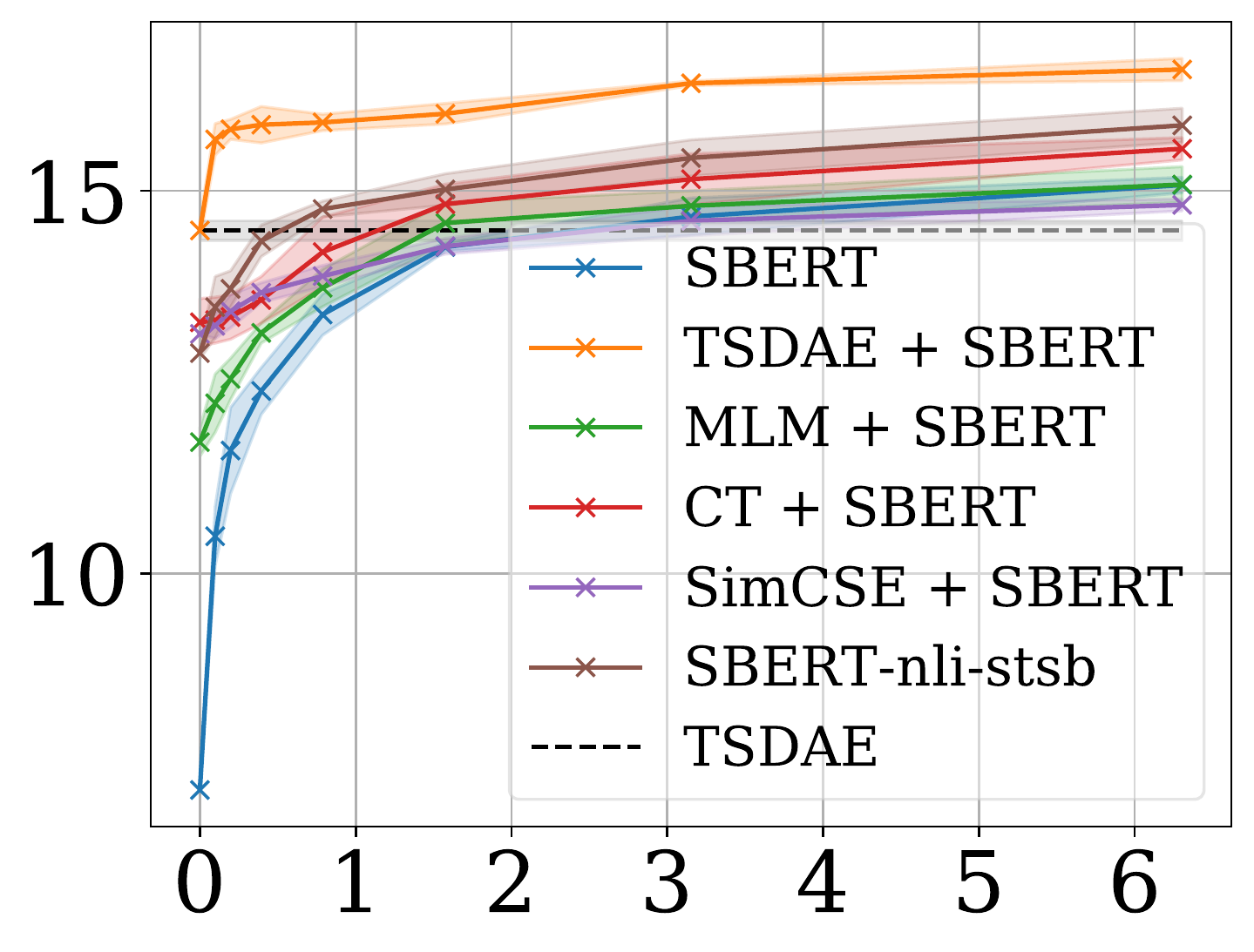}}
  \\
  \subfloat[TwitterPara]{\includegraphics[width=38mm]{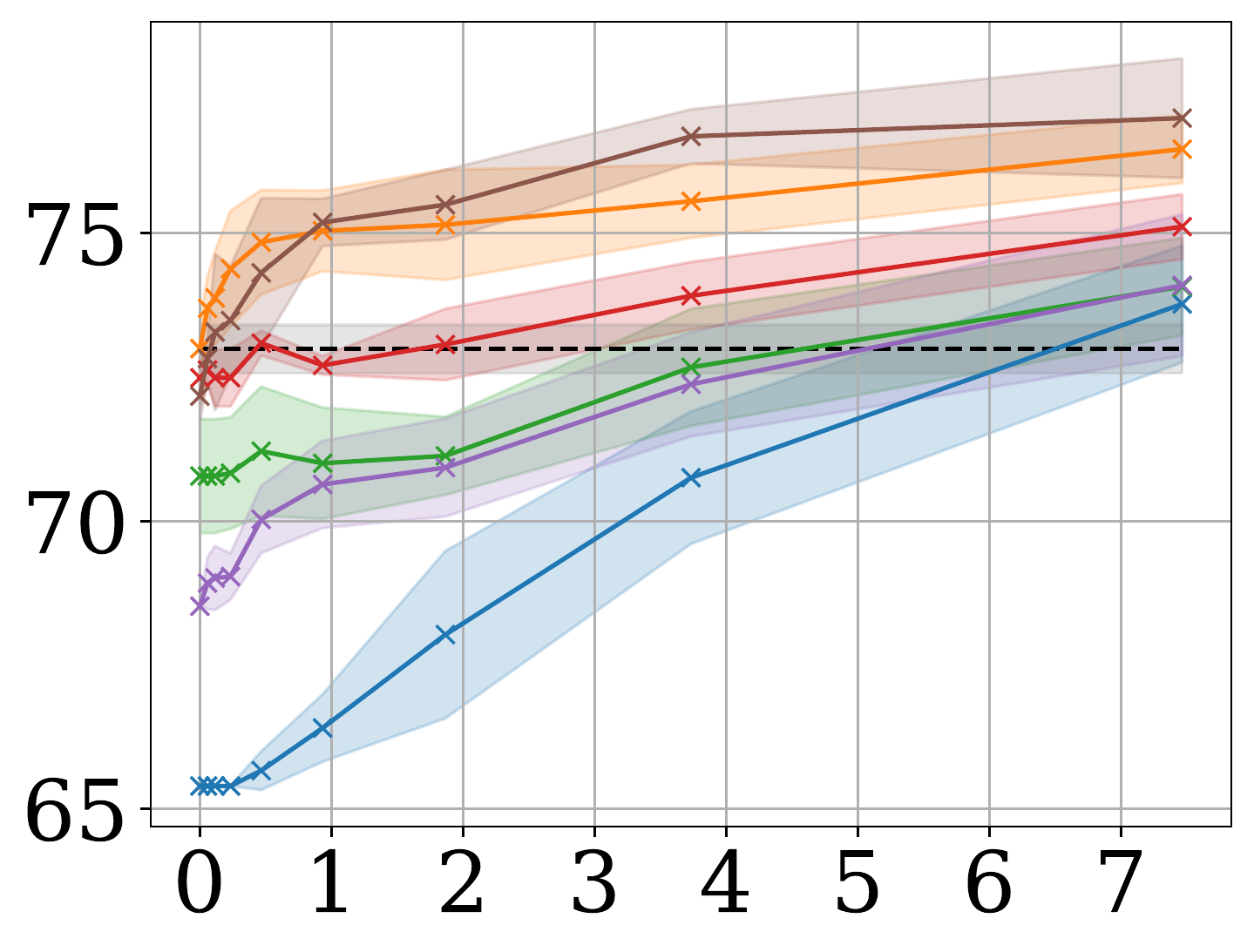}}
  \subfloat[SciDocs]{\includegraphics[width=38mm]{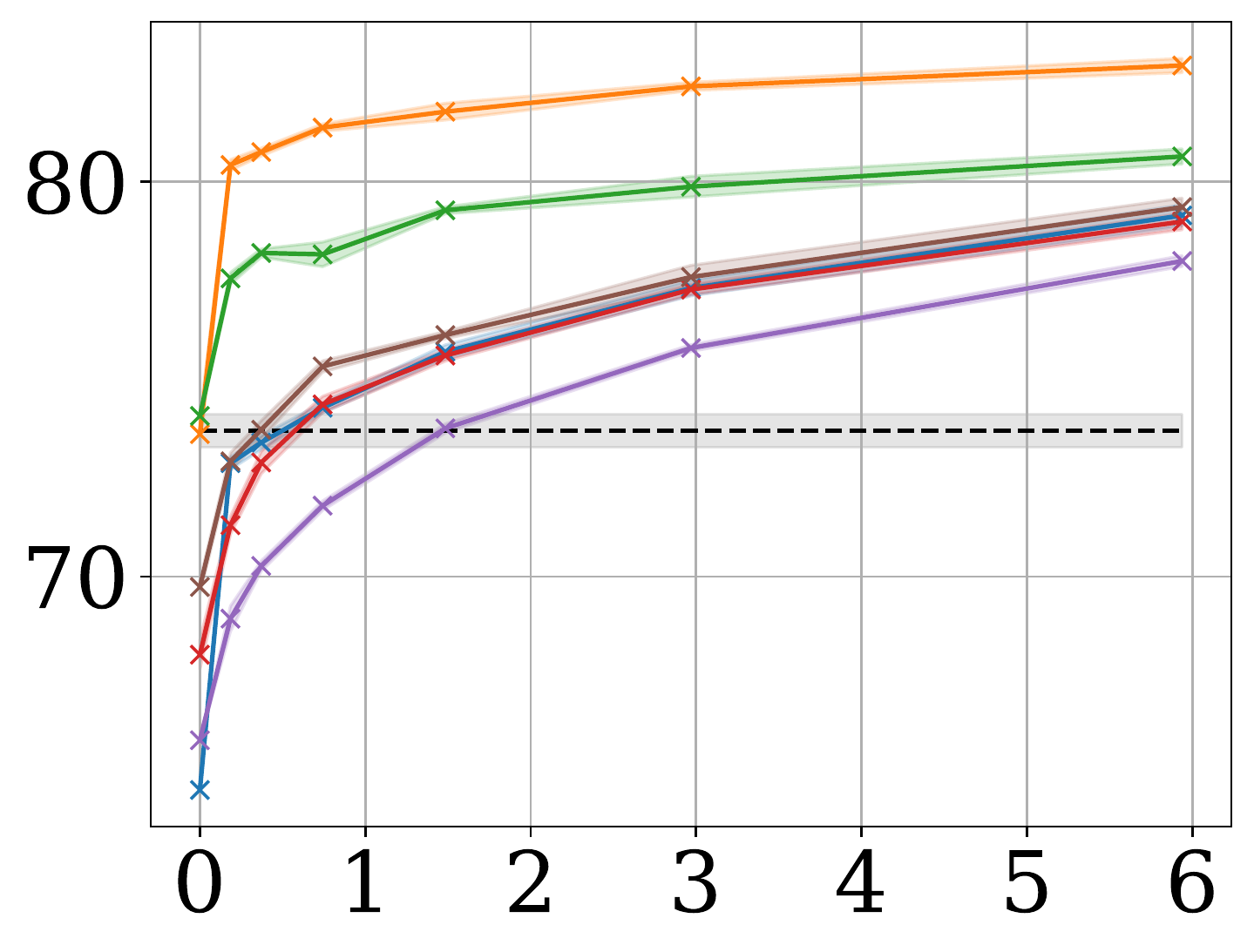}}
  \caption{Comparison of different pre-training approaches (TSDAE/MLM/CT/SimCSE+SBERT) with increasing sizes of labeled training data (in thousands). SBERT: Training from the standard \textit{BERT-base-uncased} checkpoint. TSDAE: Unsupervised baseline. Larger plots: \autoref{sec:pre_training_detailed}.}
  \label{fig:pretraining_effect}
\end{figure}

\textbf{Pre-training:} In \autoref{fig:pretraining_effect} we compare the pre-training performance of the tested approaches: We first pre-train on all available unlabeled sentences and then perform in-domain supervised training with different labeled training set sizes. Scores are reported by evaluation on the development sets. TSDAE outperforms MLM by a significant margin for all datasets except for AskUbuntu. There, MLM works slightly better. For the other datasets, TSDAE shows a clear out-performance to other pre-training strategies. The difference is quite consistent also for larger labeled training sets. We conclude, that TSDAE works well as pre-training method and can significantly improve the performance for later supervised training even for larger training datasets. CT and SimCSE don't perform well for pre-training, the results are far worse than using TSDAE/MLM or even starting from the pre-trained SBERT-nli-stsb model.

\subsection{Results on STS data}\label{sec:STS_results}

We sample sentences from Wikipedia as done by \citet{carlsson2021semantic} and train a \textit{BERT-base-uncased} model on this dataset with the different unsupervised training methods. In \autoref{tbl:sts_results}, we show the performance (Spearman's rank correlation) on the test set of the STS benchmark\footnote{In the original paper of BERT-flow, the mean pooling over the first and the last layer is used, which causes the discrepancy on the STS scores. However, for a comparable setting, as the choice of most of the previous work, we only consider the pooling over the last layer.} along with the avg.\ performance on our four domain-specific tasks. See \autoref{sec:detailed_sts} for results on other STS datasets.

\begin{table}[h]
\centering
\resizebox{6.5cm}{!}{
\begin{tabular}{|l|c|c|} 
\hline
\textbf{Method}         & \textbf{STSb}               & \textbf{Specific Tasks}                                     \\ 
\hline
\multicolumn{3}{|l|}{ \textit{Unsupervised method} }                             \\ 
\hline
TSDAE          &     66.0             &   \textbf{55.2}                                 \\
MLM     &   47.3                 &  52.9                          \\
CT             & 73.9                    & 52.4                                    \\
SimCSE         &          73.8           &   50.6                                 \\
BERT-flow      &     48.9              & 49.2                                  \\ 
\hline
\multicolumn{3}{|l|}{ \textit{Out-of-the-box supervised pre-trained models} }    \\ 
\hline
SBERT-base-nli-v2 & 83.9 & 52.0 \\
SBERT-base-nli-stsb-v2 &         \textbf{87.3}          & {52.3}                   \\
USE-large      &    80.9               &    54.7                               \\
\hline
\end{tabular}}
\caption{Performance (Spearman's rank correlation) on the STS benchmark test set. Specific tasks: Average performance from \autoref{tbl:main_results}.}
\label{tbl:sts_results}
\end{table}

We observe quite different behaviour when evaluating on STS data compared to evaluating on our domain specific tasks. On STS data, CT and SimCSE perform strongly, outperforming MLM and TSDAE by a large margin. However, when applied to domain-specific real-world tasks, TSDAE and MLM are outperforming CT and SimCSE. We think these are due to the reasons mentioned in \autoref{sec:eval}. Overall, we conclude that a strong performance on STS data is not a good indicator for good performance on domain-specific tasks.

\section{Analysis}
\label{sec:analysis}
We analyze how many training sentences are needed and if relevant content words are identified.

For all the datasets except TwitterPara, the analysis is carried out on the development set. For TwitterPara, the test set is used, as it has no development split released by the original paper. All the hyper-parameters are chosen up-front without tuning to a particular dataset.

\begin{figure}[t]
  \centering
  \subfloat[AskUbuntu]{\includegraphics[width=38mm]{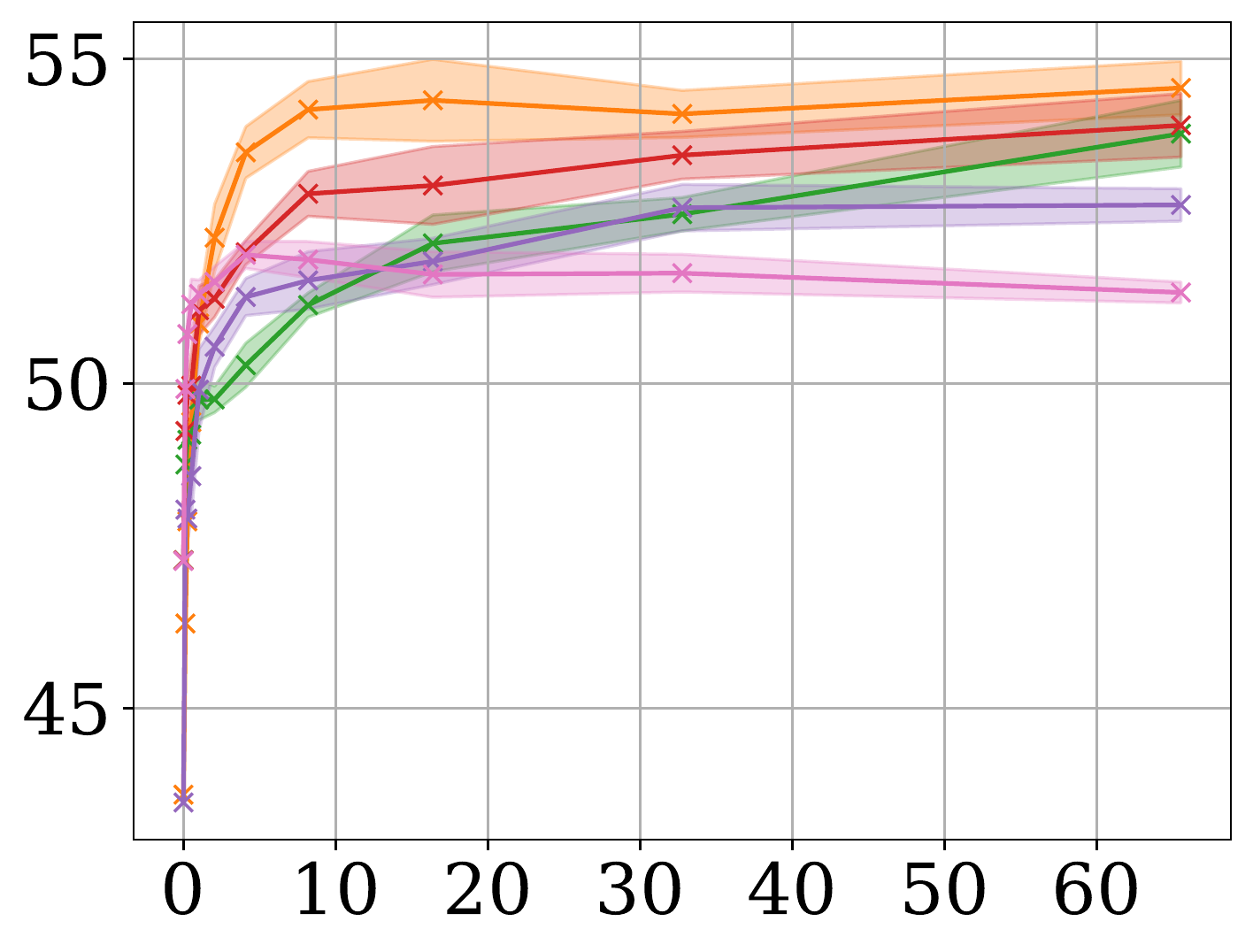}}
  \subfloat[CQADupStack]{\includegraphics[width=38mm]{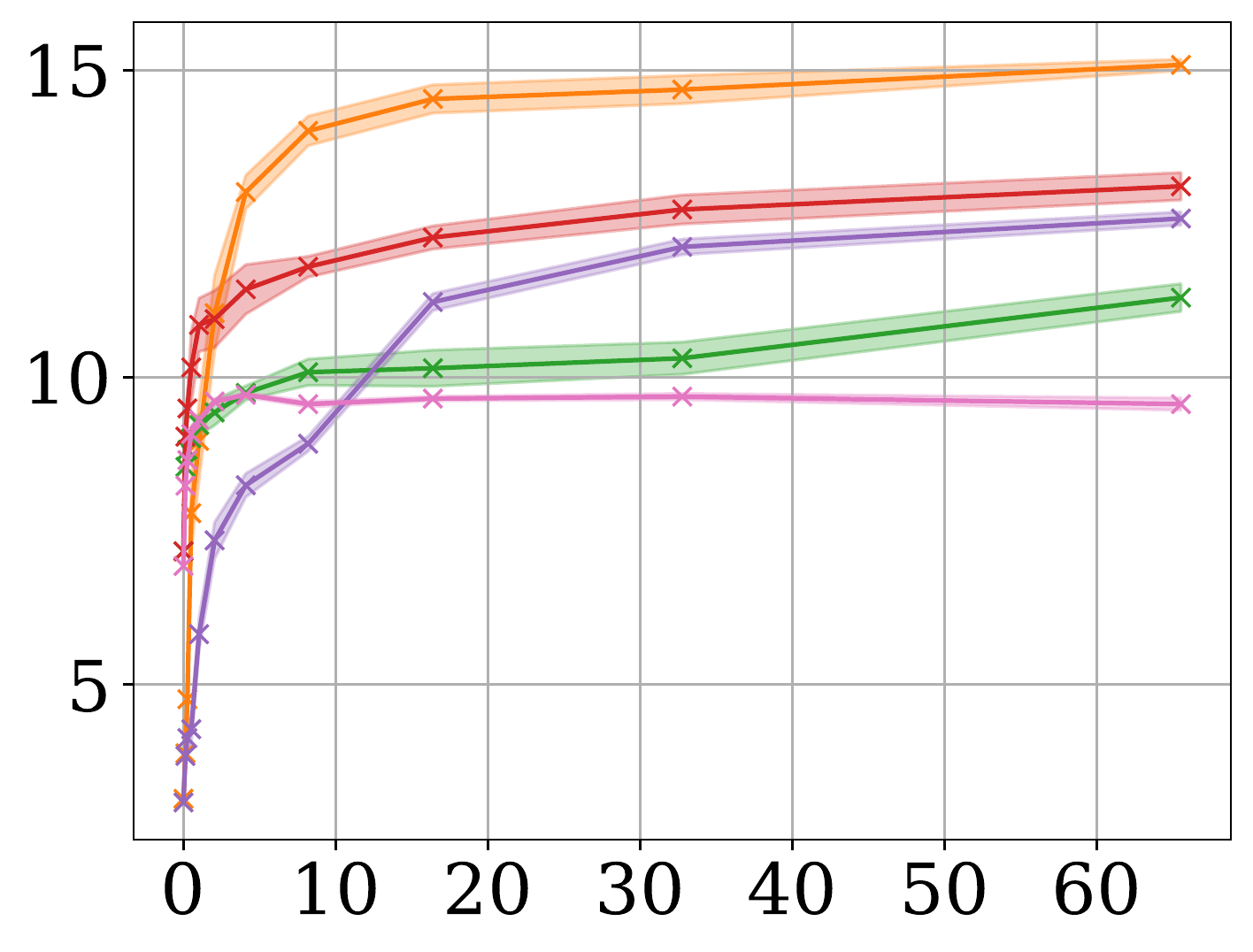}} \\
  \subfloat[TwitterPara]{\includegraphics[width=38mm]{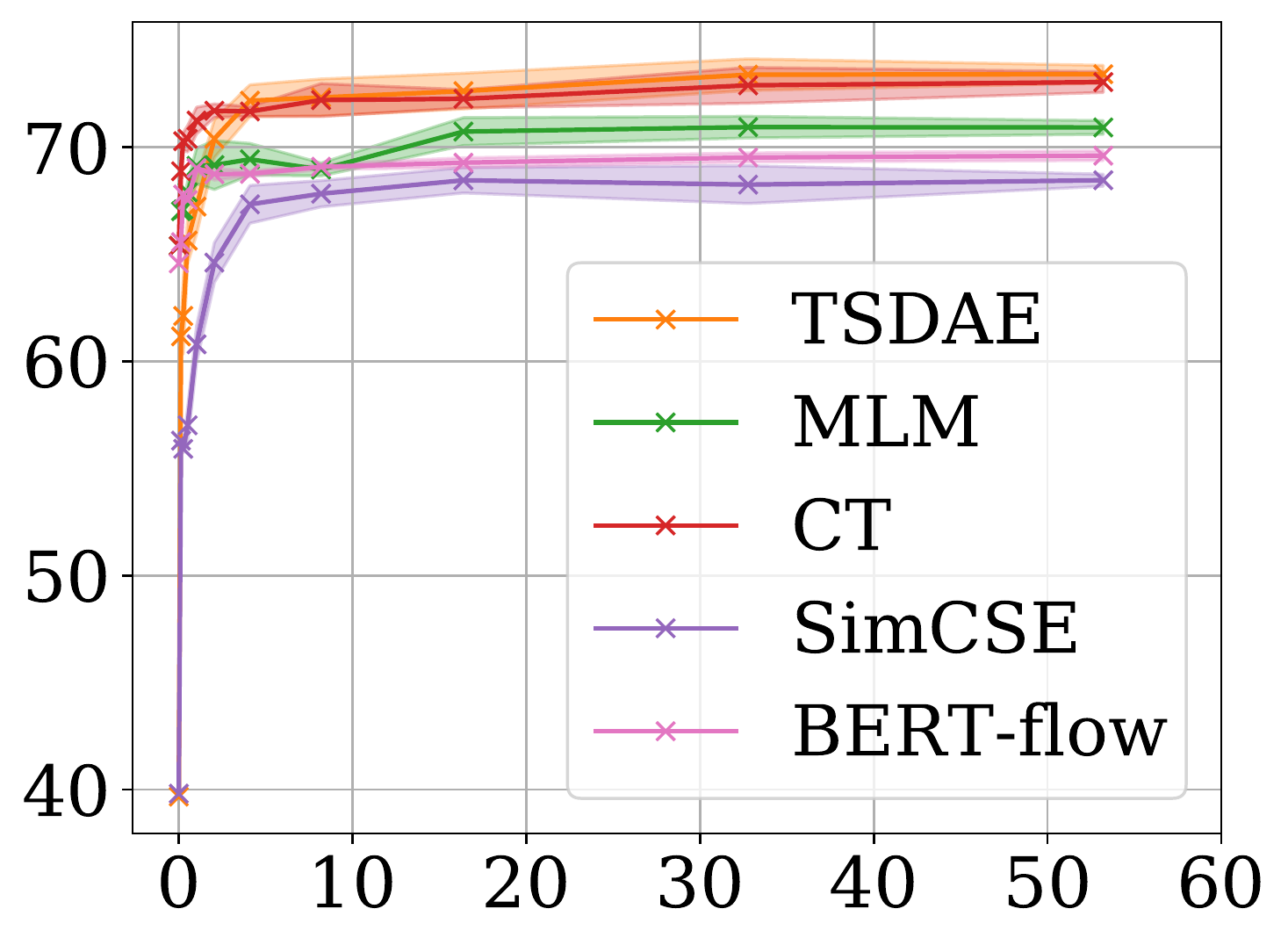}} 
  \subfloat[SciDocs]{\includegraphics[width=38mm]{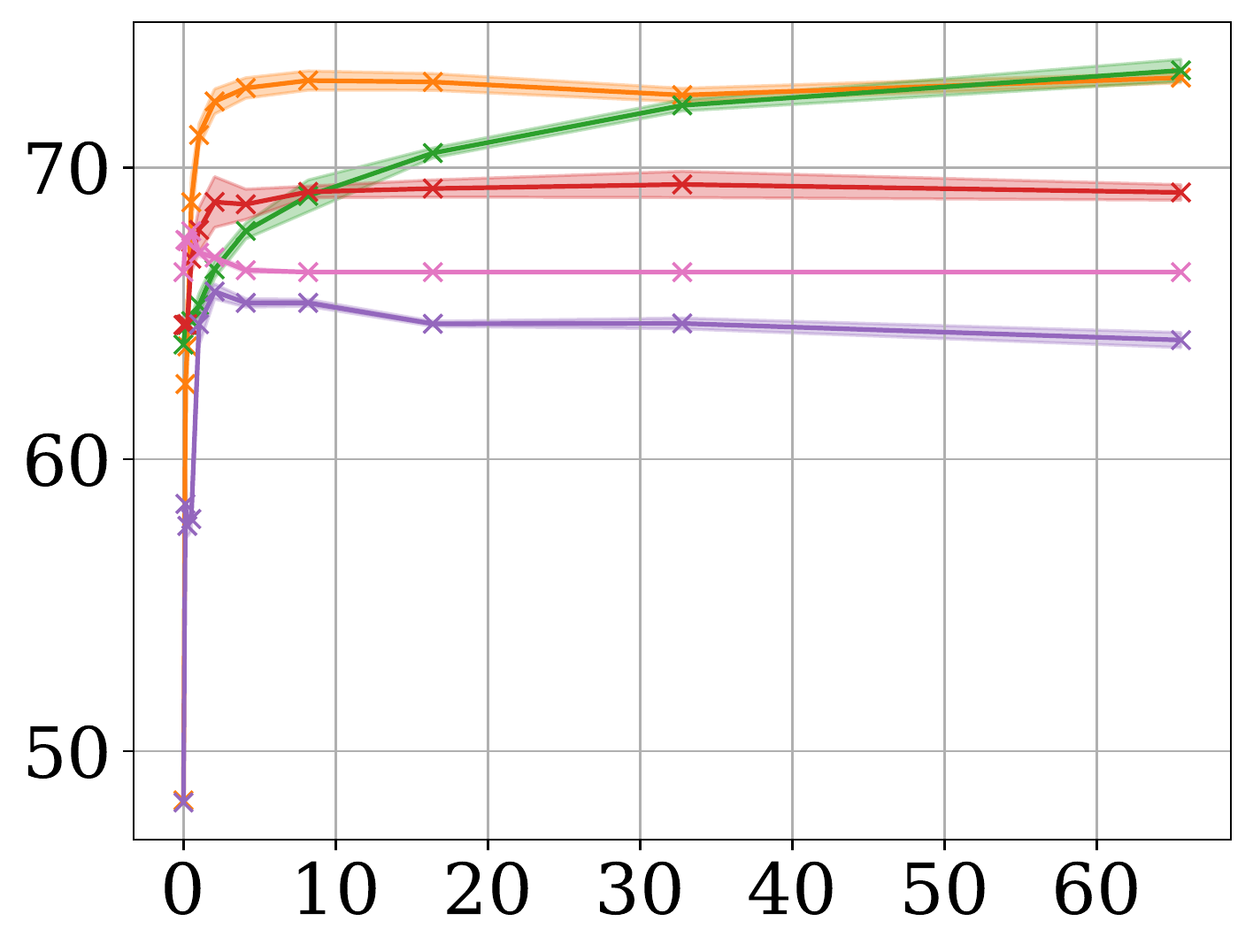}}
  \caption{The influence of the number of training sentences (in thousands) on the model performance. Larger plots: \autoref{sec:training_size_lg}.}
  \label{fig:training_size}
\end{figure}

\subsection{Influence of Corpus Size} 
In certain domains, getting a sufficiently high number of (unlabeled) sentences can be challenging. Hence, data efficiency and deriving good sentence embeddings even with little unlabeled training data can be important. 

In order to study this, we train the unsupervised approaches with different corpus sizes: Between 128 and 65,536 sentences. For each experiment, we train a bert-base-uncased model with 10 epochs up to 100k training steps. The models are evaluated at the end of each epoch and the best score on the development set is reported. 

The results are shown in \autoref{fig:training_size}. We observe that TSDAE is outperforming previous unsupervised learning methods often with as little as 1000 unlabeled sentences. With 10K unlabeled sentences, the downstream performance usually stagnates for all tested unsupervised sentence embedding methods. The only exception where more training data is helpful is for the CQADupStack task. This is expected, as the CQADupStack consists of 12 vastly different StackExchange forums, hence, requiring more unlabeled data to represent all domains well.

We conclude that comparatively little unlabeled data of $\sim$10K sentences is needed to tune pre-trained transformers to a specific domain.

\begin{figure}[t]
  \centering
  \includegraphics[width=75mm]{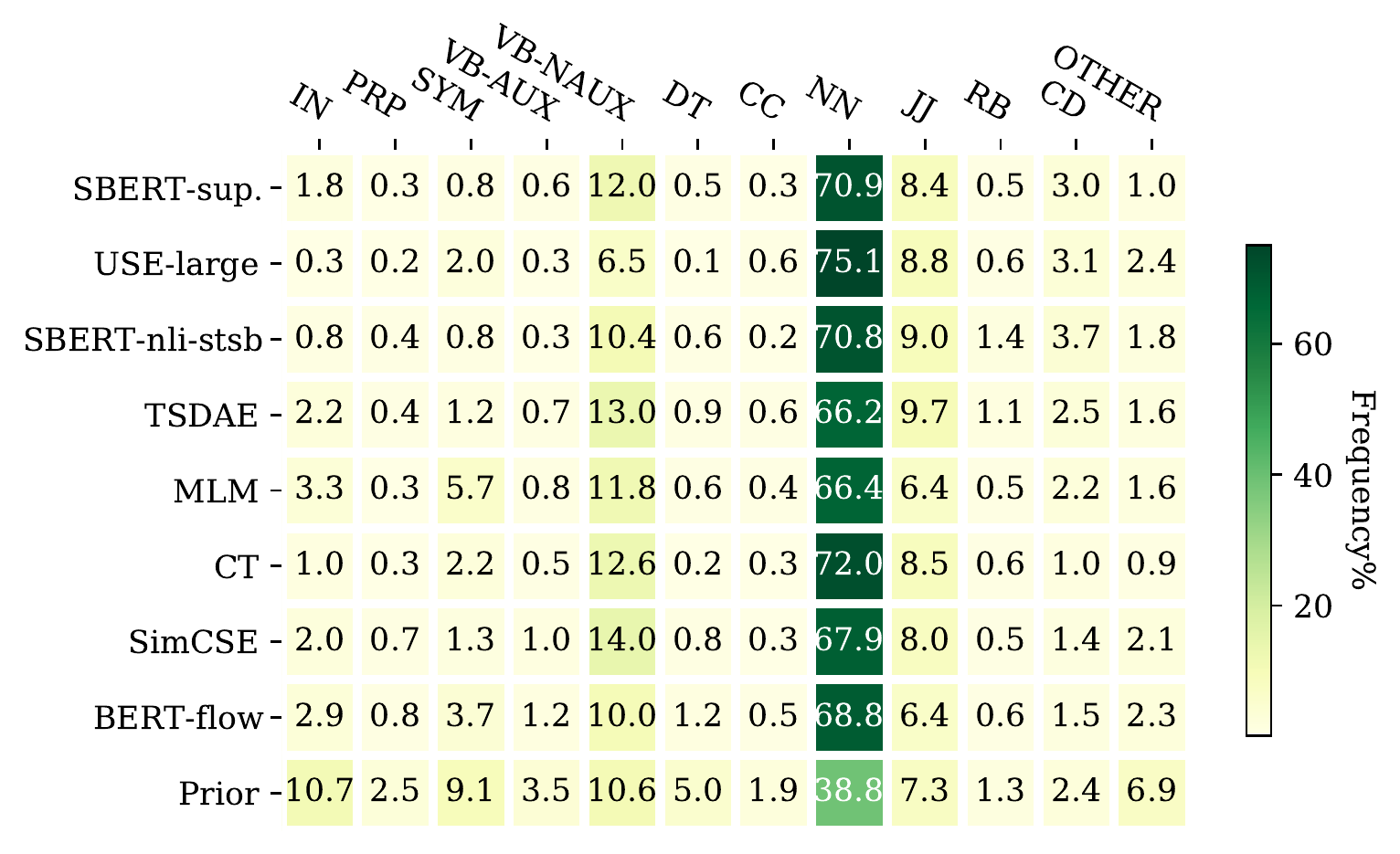}
  \caption{POS tag for the most relevant content word in a sentence, i.e.\ the word that mostly influences if a sentence pair is considered as similar. VB-AUX/-NAUX represents auxiliary/non-auxiliary verbs.}
  \label{fig:weighting_scheme_top1_overall}
\end{figure}

\subsection{Relevant Content Words}

Not all word types play an equal role in determining the semantics of a sentence. Often, nouns are the critical content words in a sentence, while e.g.\ prepositions are less important and can be add / removed from a sentences without changing the content too much.

In this section, we investigate which word types are the most relevant for the different sentence embedding methods, i.e., which words (part-of-speech tags) mainly influence if a sentence pair is perceived as similar or not. We are especially interested if we observe differences between in-domain supervised approaches (SBERT-sup.), out-of-the-box pre-trained approaches, and unsupervised approaches.  

To measure this, we select a sentence pair $(a, b)$ that is labeled as relevant and find the word that maximally reduces the cosine-similarity score for the pair (a, b):
\begin{align*}
\hat w = &\text{argmax}_w\big(\text{cossim}(a, b)-\\ &\text{min}(\text{cossim}(a\setminus w, b), \text{cossim}(a, b\setminus w))\big)
\end{align*}
among all words $w$ that appear in either a or b. Then, we record the POS tag for $\hat w$ and compute the distribution of POS tags across all sentence pairs. POS-tags are determined using CoreNLP~\citep{manning-etal-2014-stanford}.

The result averaged over the four datasets is shown in \autoref{fig:weighting_scheme_top1_overall}. For the result on each dataset, please refer to \autoref{sec:pos_tags}. Comparing the in-domain supervised model (SBERT-sup.) and the prior distribution of the POS tags, we find that nouns (NN) are by far the most relevant content words in a sentence, while function words such as prepositions (IN) and determinators (DT) have little influence on the model prediction. Surprisingly, we do not perceive significant differences between all the approaches. This is good news for the unsupervised methods (TSDAE, CT, SimCSE and BERT-flow) and show that they can learn which words types are critical in a sentence without access to labeled data. On the down side, unsupervised approaches might have issues for tasks where nouns are not the most critical content words.

\begin{table}[t]
\centering
\resizebox{7.7cm}{!}{
\begin{tabular}{|l|c|c|c|c|c|} 
\hline
\textbf{Checkpoint} & \textbf{AskU.} & \textbf{CQADup.} & \textbf{TwitterP.} & \textbf{SciDocs} & \textbf{Avg.}  \\ 
\hline
BERT-base           & 59.4/2.2     & 14.5/3.4       & 73.0/2.4         & 74.0/2.8       & 55.2/2.7     \\
Scratch             & 56.6/2.6     & 8.4/4.2        & 69.8/3.3         & 67.2/3.5       & 50.5/3.4     \\
BART-base           & 58.5/1.4     & 9.5/2.0        & 60.3/1.5         & 62.0/1.7       & 47.6/1.7     \\
T5-base             & 45.6/1.0     & 2.2/1.4        & 48.2/1.5         & 30.8/1.1       & 31.7/1.3     \\
\hline
\end{tabular}}
\caption{Test performance/training loss of TSDAE models starting from different checkpoints. The results for BERT-base are copied from \autoref{tbl:main_results}.}
\label{tbl:discussion}
\end{table}

\section{Discussion}
We mainly experiment with pre-trained Transformer encoders in this work. Besides single encoders, there are also pre-trained encoder-decoder models like BART~\cite{lewis-etal-2020-bart} and T5~\cite{DBLP:journals/jmlr/RaffelSRLNMZLL20}. However, they are already extensively pre-trained with variants of auto-encoder loss on the general domain and they are suspected of overfitting the reconstruction behavior. To verify this idea, we also further train BART-base and T5-base models with TSDAE on the 4 domain-specific datasets. The results are shown in \autoref{tbl:discussion}. We observe that BART and T5 can achieve much lower training loss (1.7 and 1.3 on average, resp.) than from scratch (3.4) or BERT (2.7), but they achieve rather bad test performance, even worse than from scratch. Compared with training from scratch (which is similar to~\citet{DBLP:conf/emnlp/ZhangWLL18}), on the other hand, we find starting from BERT can reach to a much better balance point between loss fitting and generalization. Thus, we conclude that TSDAE is more suitable to start from single encoder checkpoints, which can utilize the pre-trained knowledge while avoiding overfitting.

\section{Conclusion}
In this work, we propose a new unsupervised sentence embedding learning method based on pre-trained Transformers and sequential deoising auto-encoder (TSDAE). We evaluate TSDAE on other, recent state-of-the-art unsupervised learning on four different tasks from heterogeneous (specialized) domains in three different settings: \textit{unsupervised learning}, \textit{domain adaptation}, and \textit{pre-training}.

We observe that TSDAE performs well on the selected tasks and for the different settings, significantly outperforming other approaches.

Further, we show that the current evaluation of unsupervised sentence embedding learning approach, which is primarily done on the Semantic Textual Similarity (STS) task, is insufficient: A strong performance on STS does not correlate with a good performance on specific tasks. Many recent unsupervised approaches are not able to outperform out-of-the-box pre-trained models on the selected tasks.

\section*{Acknowledgments}
This work has been supported by the German Research Foundation (DFG) as part of the UKP-SQuARE project (grant GU 798/29-1), by the European Regional Development Fund (ERDF) and the Hessian State Chancellery – Hessian Minister of Digital Strategy and Development under the promotional reference 20005482 (TexPrax), and by the German Federal Ministry of Education and Research and the Hessian Ministry of Higher Education, Research, Science and the Arts within their joint support of the National Research Center for Applied Cybersecurity ATHENE.

\bibliography{anthology,custom}

\begin{thebibliography}{34}
\expandafter\ifx\csname natexlab\endcsname\relax\def\natexlab#1{#1}\fi

\bibitem[{Bowman et~al.(2015)Bowman, Angeli, Potts, and
  Manning}]{DBLP:conf/emnlp/BowmanAPM15}
Samuel~R. Bowman, Gabor Angeli, Christopher Potts, and Christopher~D. Manning.
  2015.
\newblock \href {https://doi.org/10.18653/v1/d15-1075} {A large annotated
  corpus for learning natural language inference}.
\newblock In \emph{Proceedings of the 2015 Conference on Empirical Methods in
  Natural Language Processing, {EMNLP} 2015, Lisbon, Portugal, September 17-21,
  2015}, pages 632--642. The Association for Computational Linguistics.

\bibitem[{Carlsson et~al.(2021)Carlsson, Sahlgren, Gogoulou, Gyllensten, and
  Hellqvist}]{carlsson2021semantic}
Fredrik Carlsson, Magnus Sahlgren, Evangelia Gogoulou, Amaru~Cuba Gyllensten,
  and Erik~Ylip{\"a}{\"a} Hellqvist. 2021.
\newblock \href {https://openreview.net/forum?id=Ov_sMNau-PF} {Semantic
  re-tuning with contrastive tension}.
\newblock In \emph{International Conference on Learning Representations, {ICLR}
  2021, Vienna, Austria, May 3-7, 2021}, pages 1--21. International Conference
  on Learning Representations.

\bibitem[{Cer et~al.(2017)Cer, Diab, Agirre, Lopez-Gazpio, and
  Specia}]{cer-etal-2017-semeval}
Daniel Cer, Mona Diab, Eneko Agirre, I{\~n}igo Lopez-Gazpio, and Lucia Specia.
  2017.
\newblock \href {https://doi.org/10.18653/v1/S17-2001} {{S}em{E}val-2017 task
  1: Semantic textual similarity multilingual and crosslingual focused
  evaluation}.
\newblock In \emph{Proceedings of the 11th International Workshop on Semantic
  Evaluation ({S}em{E}val-2017)}, pages 1--14, Vancouver, Canada. Association
  for Computational Linguistics.

\bibitem[{Cer et~al.(2018)Cer, Yang, Kong, Hua, Limtiaco, John, Constant,
  Guajardo{-}Cespedes, Yuan, Tar, Strope, and
  Kurzweil}]{DBLP:conf/emnlp/CerYKHLJCGYTSK18}
Daniel Cer, Yinfei Yang, Sheng{-}yi Kong, Nan Hua, Nicole Limtiaco, Rhomni~St.
  John, Noah Constant, Mario Guajardo{-}Cespedes, Steve Yuan, Chris Tar, Brian
  Strope, and Ray Kurzweil. 2018.
\newblock \href {https://doi.org/10.18653/v1/d18-2029} {Universal sentence
  encoder for english}.
\newblock In \emph{Proceedings of the 2018 Conference on Empirical Methods in
  Natural Language Processing, {EMNLP} 2018: System Demonstrations, Brussels,
  Belgium, October 31 - November 4, 2018}, pages 169--174. Association for
  Computational Linguistics.

\bibitem[{Chen et~al.(2020)Chen, Kornblith, Norouzi, and
  Hinton}]{DBLP:conf/icml/ChenK0H20}
Ting Chen, Simon Kornblith, Mohammad Norouzi, and Geoffrey~E. Hinton. 2020.
\newblock \href {http://proceedings.mlr.press/v119/chen20j.html} {A simple
  framework for contrastive learning of visual representations}.
\newblock In \emph{Proceedings of the 37th International Conference on Machine
  Learning, {ICML} 2020, 13-18 July 2020, Virtual Event}, volume 119 of
  \emph{Proceedings of Machine Learning Research}, pages 1597--1607. {PMLR}.

\bibitem[{Cohan et~al.(2020)Cohan, Feldman, Beltagy, Downey, and
  Weld}]{DBLP:conf/acl/CohanFBDW20}
Arman Cohan, Sergey Feldman, Iz~Beltagy, Doug Downey, and Daniel~S. Weld. 2020.
\newblock \href {https://doi.org/10.18653/v1/2020.acl-main.207} {{SPECTER:}
  document-level representation learning using citation-informed transformers}.
\newblock In \emph{Proceedings of the 58th Annual Meeting of the Association
  for Computational Linguistics, {ACL} 2020, Online, July 5-10, 2020}, pages
  2270--2282. Association for Computational Linguistics.

\bibitem[{Conneau et~al.(2017)Conneau, Kiela, Schwenk, Barrault, and
  Bordes}]{conneau-EtAl:2017:EMNLP2017}
Alexis Conneau, Douwe Kiela, Holger Schwenk, Lo\"{i}c Barrault, and Antoine
  Bordes. 2017.
\newblock \href {https://www.aclweb.org/anthology/D17-1070} {Supervised
  learning of universal sentence representations from natural language
  inference data}.
\newblock In \emph{Proceedings of the 2017 Conference on Empirical Methods in
  Natural Language Processing}, pages 670--680, Copenhagen, Denmark.
  Association for Computational Linguistics.

\bibitem[{Devlin et~al.(2019)Devlin, Chang, Lee, and
  Toutanova}]{DBLP:conf/naacl/DevlinCLT19}
Jacob Devlin, Ming{-}Wei Chang, Kenton Lee, and Kristina Toutanova. 2019.
\newblock \href {https://doi.org/10.18653/v1/n19-1423} {{BERT:} pre-training of
  deep bidirectional transformers for language understanding}.
\newblock In \emph{Proceedings of the 2019 Conference of the North American
  Chapter of the Association for Computational Linguistics: Human Language
  Technologies, {NAACL-HLT} 2019, Minneapolis, MN, USA, June 2-7, 2019, Volume
  1 (Long and Short Papers)}, pages 4171--4186. Association for Computational
  Linguistics.

\bibitem[{Gao et~al.(2021)Gao, Yao, and Chen}]{gao2021simcse}
Tianyu Gao, Xingcheng Yao, and Danqi Chen. 2021.
\newblock \href {https://arxiv.org/abs/2104.08821} {{SimCSE}: Simple
  contrastive learning of sentence embeddings}.
\newblock \emph{arXiv preprint arXiv:2104.08821}.

\bibitem[{Giorgi et~al.(2021)Giorgi, Nitski, Wang, and
  Bader}]{DBLP:journals/corr/abs-2006-03659}
John Giorgi, Osvald Nitski, Bo~Wang, and Gary Bader. 2021.
\newblock \href {https://doi.org/10.18653/v1/2021.acl-long.72} {{D}e{CLUTR}:
  Deep contrastive learning for unsupervised textual representations}.
\newblock In \emph{Proceedings of the 59th Annual Meeting of the Association
  for Computational Linguistics and the 11th International Joint Conference on
  Natural Language Processing (Volume 1: Long Papers)}, pages 879--895, Online.
  Association for Computational Linguistics.

\bibitem[{Goodfellow et~al.(2016)Goodfellow, Bengio, and
  Courville}]{Goodfellow-et-al-2016}
Ian Goodfellow, Yoshua Bengio, and Aaron Courville. 2016.
\newblock \emph{Deep Learning}.
\newblock MIT Press.
\newblock \url{http://www.deeplearningbook.org}.

\bibitem[{Hadsell et~al.(2006)Hadsell, Chopra, and
  LeCun}]{DBLP:conf/cvpr/HadsellCL06}
Raia Hadsell, Sumit Chopra, and Yann LeCun. 2006.
\newblock \href {https://doi.org/10.1109/CVPR.2006.100} {Dimensionality
  reduction by learning an invariant mapping}.
\newblock In \emph{2006 {IEEE} Computer Society Conference on Computer Vision
  and Pattern Recognition {(CVPR} 2006), 17-22 June 2006, New York, NY, {USA}},
  pages 1735--1742. {IEEE} Computer Society.

\bibitem[{Henderson et~al.(2017)Henderson, Al{-}Rfou, Strope, Sung,
  Luk{\'{a}}cs, Guo, Kumar, Miklos, and
  Kurzweil}]{DBLP:journals/corr/HendersonASSLGK17}
Matthew~L. Henderson, Rami Al{-}Rfou, Brian Strope, Yun{-}Hsuan Sung,
  L{\'{a}}szl{\'{o}} Luk{\'{a}}cs, Ruiqi Guo, Sanjiv Kumar, Balint Miklos, and
  Ray Kurzweil. 2017.
\newblock \href {http://arxiv.org/abs/1705.00652} {{Efficient Natural Language
  Response Suggestion for Smart Reply}}.
\newblock \emph{arXiv preprint arXiv:1705.00652}.

\bibitem[{Hill et~al.(2016)Hill, Cho, and Korhonen}]{DBLP:conf/naacl/HillCK16}
Felix Hill, Kyunghyun Cho, and Anna Korhonen. 2016.
\newblock \href {https://doi.org/10.18653/v1/n16-1162} {Learning distributed
  representations of sentences from unlabelled data}.
\newblock In \emph{{NAACL} {HLT} 2016, The 2016 Conference of the North
  American Chapter of the Association for Computational Linguistics: Human
  Language Technologies, San Diego California, USA, June 12-17, 2016}, pages
  1367--1377. The Association for Computational Linguistics.

\bibitem[{Hoogeveen et~al.(2015)Hoogeveen, Verspoor, and
  Baldwin}]{DBLP:conf/adcs/HoogeveenVB15}
Doris Hoogeveen, Karin~M. Verspoor, and Timothy Baldwin. 2015.
\newblock \href {https://doi.org/10.1145/2838931.2838934} {Cqadupstack: {A}
  benchmark data set for community question-answering research}.
\newblock In \emph{Proceedings of the 20th Australasian Document Computing
  Symposium, {ADCS} 2015, Parramatta, NSW, Australia, December 8-9, 2015},
  pages 3:1--3:8. {ACM}.

\bibitem[{Kingma and Dhariwal(2018)}]{DBLP:conf/nips/KingmaD18}
Diederik~P. Kingma and Prafulla Dhariwal. 2018.
\newblock \href
  {https://proceedings.neurips.cc/paper/2018/hash/d139db6a236200b21cc7f752979132d0-Abstract.html}
  {Glow: Generative flow with invertible 1x1 convolutions}.
\newblock In \emph{Advances in Neural Information Processing Systems 31: Annual
  Conference on Neural Information Processing Systems 2018, NeurIPS 2018,
  December 3-8, 2018, Montr{\'{e}}al, Canada}, pages 10236--10245.

\bibitem[{Lan et~al.(2017)Lan, Qiu, He, and Xu}]{DBLP:conf/emnlp/LanQHX17}
Wuwei Lan, Siyu Qiu, Hua He, and Wei Xu. 2017.
\newblock \href {https://doi.org/10.18653/v1/d17-1126} {A continuously growing
  dataset of sentential paraphrases}.
\newblock In \emph{Proceedings of the 2017 Conference on Empirical Methods in
  Natural Language Processing, {EMNLP} 2017, Copenhagen, Denmark, September
  9-11, 2017}, pages 1224--1234. Association for Computational Linguistics.

\bibitem[{Lei et~al.(2016)Lei, Joshi, Barzilay, Jaakkola, Tymoshenko,
  Moschitti, and M{\`{a}}rquez}]{DBLP:conf/naacl/LeiJBJTMM16}
Tao Lei, Hrishikesh Joshi, Regina Barzilay, Tommi~S. Jaakkola, Kateryna
  Tymoshenko, Alessandro Moschitti, and Llu{\'{\i}}s M{\`{a}}rquez. 2016.
\newblock \href {https://doi.org/10.18653/v1/n16-1153} {Semi-supervised
  question retrieval with gated convolutions}.
\newblock In \emph{{NAACL} {HLT} 2016, The 2016 Conference of the North
  American Chapter of the Association for Computational Linguistics: Human
  Language Technologies, San Diego California, USA, June 12-17, 2016}, pages
  1279--1289. The Association for Computational Linguistics.

\bibitem[{Lewis et~al.(2020)Lewis, Liu, Goyal, Ghazvininejad, Mohamed, Levy,
  Stoyanov, and Zettlemoyer}]{lewis-etal-2020-bart}
Mike Lewis, Yinhan Liu, Naman Goyal, Marjan Ghazvininejad, Abdelrahman Mohamed,
  Omer Levy, Veselin Stoyanov, and Luke Zettlemoyer. 2020.
\newblock \href {https://doi.org/10.18653/v1/2020.acl-main.703} {{BART}:
  Denoising sequence-to-sequence pre-training for natural language generation,
  translation, and comprehension}.
\newblock In \emph{Proceedings of the 58th Annual Meeting of the Association
  for Computational Linguistics}, pages 7871--7880, Online. Association for
  Computational Linguistics.

\bibitem[{Li et~al.(2020)Li, Zhou, He, Wang, Yang, and
  Li}]{DBLP:conf/emnlp/LiZHWYL20}
Bohan Li, Hao Zhou, Junxian He, Mingxuan Wang, Yiming Yang, and Lei Li. 2020.
\newblock \href {https://doi.org/10.18653/v1/2020.emnlp-main.733} {On the
  sentence embeddings from pre-trained language models}.
\newblock In \emph{Proceedings of the 2020 Conference on Empirical Methods in
  Natural Language Processing, {EMNLP} 2020, Online, November 16-20, 2020},
  pages 9119--9130. Association for Computational Linguistics.

\bibitem[{Manning et~al.(2014)Manning, Surdeanu, Bauer, Finkel, Bethard, and
  McClosky}]{manning-etal-2014-stanford}
Christopher Manning, Mihai Surdeanu, John Bauer, Jenny Finkel, Steven Bethard,
  and David McClosky. 2014.
\newblock \href {https://doi.org/10.3115/v1/P14-5010} {The {S}tanford
  {C}ore{NLP} natural language processing toolkit}.
\newblock In \emph{Proceedings of 52nd Annual Meeting of the Association for
  Computational Linguistics: System Demonstrations}, pages 55--60, Baltimore,
  Maryland. Association for Computational Linguistics.

\bibitem[{Pagliardini et~al.(2018)Pagliardini, Gupta, and
  Jaggi}]{DBLP:conf/naacl/PagliardiniGJ18}
Matteo Pagliardini, Prakhar Gupta, and Martin Jaggi. 2018.
\newblock \href {https://doi.org/10.18653/v1/n18-1049} {Unsupervised learning
  of sentence embeddings using compositional n-gram features}.
\newblock In \emph{Proceedings of the 2018 Conference of the North American
  Chapter of the Association for Computational Linguistics: Human Language
  Technologies, {NAACL-HLT} 2018, New Orleans, Louisiana, USA, June 1-6, 2018,
  Volume 1 (Long Papers)}, pages 528--540. Association for Computational
  Linguistics.

\bibitem[{Pennington et~al.(2014)Pennington, Socher, and
  Manning}]{pennington2014glove}
Jeffrey Pennington, Richard Socher, and Christopher~D. Manning. 2014.
\newblock \href {http://www.aclweb.org/anthology/D14-1162} {Glove: Global
  vectors for word representation}.
\newblock In \emph{Empirical Methods in Natural Language Processing (EMNLP)},
  pages 1532--1543.

\bibitem[{Raffel et~al.(2020)Raffel, Shazeer, Roberts, Lee, Narang, Matena,
  Zhou, Li, and Liu}]{DBLP:journals/jmlr/RaffelSRLNMZLL20}
Colin Raffel, Noam Shazeer, Adam Roberts, Katherine Lee, Sharan Narang, Michael
  Matena, Yanqi Zhou, Wei Li, and Peter~J. Liu. 2020.
\newblock \href {http://jmlr.org/papers/v21/20-074.html} {Exploring the limits
  of transfer learning with a unified text-to-text transformer}.
\newblock \emph{J. Mach. Learn. Res.}, 21:140:1--140:67.

\bibitem[{Reimers et~al.(2016)Reimers, Beyer, and
  Gurevych}]{DBLP:conf/coling/ReimersBG16}
Nils Reimers, Philip Beyer, and Iryna Gurevych. 2016.
\newblock \href {https://www.aclweb.org/anthology/C16-1009/} {Task-oriented
  intrinsic evaluation of semantic textual similarity}.
\newblock In \emph{{COLING} 2016, 26th International Conference on
  Computational Linguistics, Proceedings of the Conference: Technical Papers,
  December 11-16, 2016, Osaka, Japan}, pages 87--96. {ACL}.

\bibitem[{Reimers and Gurevych(2019)}]{DBLP:conf/emnlp/ReimersG19}
Nils Reimers and Iryna Gurevych. 2019.
\newblock \href {https://doi.org/10.18653/v1/D19-1410} {Sentence-bert: Sentence
  embeddings using siamese bert-networks}.
\newblock In \emph{Proceedings of the 2019 Conference on Empirical Methods in
  Natural Language Processing and the 9th International Joint Conference on
  Natural Language Processing, {EMNLP-IJCNLP} 2019, Hong Kong, China, November
  3-7, 2019}, pages 3980--3990. Association for Computational Linguistics.

\bibitem[{Robertson et~al.(1994)Robertson, Walker, Jones, Hancock{-}Beaulieu,
  and Gatford}]{DBLP:conf/trec/RobertsonWJHG94}
Stephen~E. Robertson, Steve Walker, Susan Jones, Micheline Hancock{-}Beaulieu,
  and Mike Gatford. 1994.
\newblock \href {http://trec.nist.gov/pubs/trec3/papers/city.ps.gz} {Okapi at
  {TREC-3}}.
\newblock In \emph{Proceedings of The Third Text REtrieval Conference, {TREC}
  1994, Gaithersburg, Maryland, USA, November 2-4, 1994}, volume 500-225 of
  \emph{{NIST} Special Publication}, pages 109--126. National Institute of
  Standards and Technology {(NIST)}.

\bibitem[{Su et~al.(2021)Su, Cao, Liu, and Ou}]{su2021whitening}
Jianlin Su, Jiarun Cao, Weijie Liu, and Yangyiwen Ou. 2021.
\newblock \href {https://arxiv.org/abs/2103.15316} {{Whitening Sentence
  Representations for Better Semantics and Faster Retrieval}}.
\newblock \emph{arXiv preprint arXiv:2103.15316}.

\bibitem[{Vaswani et~al.(2017)Vaswani, Shazeer, Parmar, Uszkoreit, Jones,
  Gomez, Kaiser, and Polosukhin}]{DBLP:conf/nips/VaswaniSPUJGKP17}
Ashish Vaswani, Noam Shazeer, Niki Parmar, Jakob Uszkoreit, Llion Jones,
  Aidan~N. Gomez, Lukasz Kaiser, and Illia Polosukhin. 2017.
\newblock \href
  {https://proceedings.neurips.cc/paper/2017/hash/3f5ee243547dee91fbd053c1c4a845aa-Abstract.html}
  {Attention is all you need}.
\newblock In \emph{Advances in Neural Information Processing Systems 30: Annual
  Conference on Neural Information Processing Systems 2017, December 4-9, 2017,
  Long Beach, CA, {USA}}, pages 5998--6008.

\bibitem[{Vincent et~al.(2010)Vincent, Larochelle, Lajoie, Bengio, and
  Manzagol}]{DBLP:journals/jmlr/VincentLLBM10}
Pascal Vincent, Hugo Larochelle, Isabelle Lajoie, Yoshua Bengio, and
  Pierre{-}Antoine Manzagol. 2010.
\newblock \href {http://portal.acm.org/citation.cfm?id=1953039} {Stacked
  denoising autoencoders: Learning useful representations in a deep network
  with a local denoising criterion}.
\newblock \emph{J. Mach. Learn. Res.}, 11:3371--3408.

\bibitem[{Williams et~al.(2018)Williams, Nangia, and
  Bowman}]{DBLP:conf/naacl/WilliamsNB18}
Adina Williams, Nikita Nangia, and Samuel~R. Bowman. 2018.
\newblock \href {https://doi.org/10.18653/v1/n18-1101} {A broad-coverage
  challenge corpus for sentence understanding through inference}.
\newblock In \emph{Proceedings of the 2018 Conference of the North American
  Chapter of the Association for Computational Linguistics: Human Language
  Technologies, {NAACL-HLT} 2018, New Orleans, Louisiana, USA, June 1-6, 2018,
  Volume 1 (Long Papers)}, pages 1112--1122. Association for Computational
  Linguistics.

\bibitem[{Xu et~al.(2015)Xu, Callison{-}Burch, and
  Dolan}]{DBLP:conf/semeval/XuCD15}
Wei Xu, Chris Callison{-}Burch, and Bill Dolan. 2015.
\newblock \href {https://doi.org/10.18653/v1/s15-2001} {Semeval-2015 task 1:
  Paraphrase and semantic similarity in twitter {(PIT)}}.
\newblock In \emph{Proceedings of the 9th International Workshop on Semantic
  Evaluation, SemEval@NAACL-HLT 2015, Denver, Colorado, USA, June 4-5, 2015},
  pages 1--11. The Association for Computer Linguistics.

\bibitem[{Yang et~al.(2020)Yang, Cer, Ahmad, Guo, Law, Constant, {\'{A}}brego,
  Yuan, Tar, Sung, Strope, and Kurzweil}]{DBLP:conf/acl/YangCAGLCAYTSSK20}
Yinfei Yang, Daniel Cer, Amin Ahmad, Mandy Guo, Jax Law, Noah Constant,
  Gustavo~Hern{\'{a}}ndez {\'{A}}brego, Steve Yuan, Chris Tar, Yun{-}Hsuan
  Sung, Brian Strope, and Ray Kurzweil. 2020.
\newblock \href {https://doi.org/10.18653/v1/2020.acl-demos.12} {Multilingual
  universal sentence encoder for semantic retrieval}.
\newblock In \emph{Proceedings of the 58th Annual Meeting of the Association
  for Computational Linguistics: System Demonstrations, {ACL} 2020, Online,
  July 5-10, 2020}, pages 87--94. Association for Computational Linguistics.

\bibitem[{Zhang et~al.(2018)Zhang, Wu, Li, and Li}]{DBLP:conf/emnlp/ZhangWLL18}
Minghua Zhang, Yunfang Wu, Weikang Li, and Wei Li. 2018.
\newblock \href {https://aclanthology.org/D18-1481/} {Learning universal
  sentence representations with mean-max attention autoencoder}.
\newblock In \emph{Proceedings of the 2018 Conference on Empirical Methods in
  Natural Language Processing, Brussels, Belgium, October 31 - November 4,
  2018}, pages 4514--4523. Association for Computational Linguistics.

\end{thebibliography}
\bibliographystyle{acl_natbib}

\onecolumn
\appendix

\section{Optimal Configuration of TSDAE}
\label{sec:TSDAE_config}
To obtain the optimal configuration, we compare TSDAE models trained and evaluated on the general domain without bias towards any specific domain. The greedy search is applied by sequentially finding the best (1) noise type and ratio (2) pooling method and (3) weight tying scheme. Similar to the choice of CT and BERT-flow, we train the models on the combination of SNLI and MultiNLI without labels and evaluate the models on the STS benchmark with the metric of Spearman rank correlation. The maximum number of training steps is 30K and the models are evaluated every 1.5K training steps, reporting the best validation performance. Scores are obtained by calculating the average over 5 random seeds.

We first compare the scores of different noise types, fixing the noise ratio as 0.3 (i.e. 30\% tokens are influenced) and the pooling method as CLS pooling. The results are show in \autoref{tbl:noise_type}. This indicates deletion is the best noise type. We then tune the noise ratio of the deletion noise and the results are shown in \autoref{tbl:noise_ratio}. This indicates 0.6 is the best noise ratio. 

\begin{table}[H]
\centering
\resizebox{8cm}{!}{
\begin{tabular}{|l|c|c|c|c|c|} 
\hline
 \textbf{Type} & Delete         & Swap  & Mask  & Replace & Add    \\ 
\hline
 \textbf{Score} & \textbf{78.33} & 76.85 & 76.56 & 74.01~  & 72.65  \\
\hline
\end{tabular}}
\caption{Results with different noise types}
\label{tbl:noise_type}
\end{table}


\begin{table}[H]
\centering
\resizebox{12cm}{!}{
\begin{tabular}{|l|c|c|c|c|c|c|c|c|c|} 
\hline
\textbf{Ratio} & 0.1   & 0.2   & 0.3   & 0.4   & 0.5   & 0.6   & 0.7   & 0.8   & 0.9    \\ 
\hline
\textbf{Score} & 77.81 & 77.70 & 77.75 & 78.02 & 78.25 & \textbf{78.77} & 78.19 & 77.69 & 75.67  \\
\hline
\end{tabular}}
\caption{Results with different noise ratio.}
\label{tbl:noise_ratio}
\end{table}

We then compare different pooling methods with the best setting so far. The results are shown in Table~\ref{tbl:pooling}. Since there is little difference between CLS and mean pooling and mean pooling loses the position information, the CLS pooling is chosen. Finally, we find that tying the encoder and the decoder can further improve the validation score to 79.15.

\begin{table}[H]
\centering
\resizebox{5cm}{!}{
\begin{tabular}{|l|c|c|c|} 
\hline
 \textbf{Method} & CLS   & Mean           & Max     \\ 
\hline
 \textbf{Score} & 78.77 & \textbf{78.84} & 78.17~  \\
\hline
\end{tabular}}
\caption{Results with different pooling methods.}
\label{tbl:pooling}
\end{table}

\section{Experiment Settings}
\label{sec:experiment_settings}
We implement TSDAE, CT and BERT-flow based on Pytorch and Huggingface's Transformers\footnote{\url{https://github.com/huggingface/transformers}} (version number: v3.1.0). For these three unsupervised methods, following the original papers, the number of training steps is 100K; the batch size is 8; the optimizers are AdamW, RMSProp and AdamW, respectively; the initial learning rates are 3e-5, 1e-5 and 1e-6, resp. The weight decay for BERT-flow is 0.01. The learning rate for CT follows a segmented-constant scheduling scheme: 1e-5 for step 1 to 500; 8e-6 for step 501 to 1000; 6e-6 for step 1001 to 1500; 4e-6 for step 1501 to 2000; 2e-6 for others. The pooling method for CT and BERT-flow is both mean pooling. Since CT trains two independent encoders and we find the second encoder has better performance, we use the second encoder for evaluation. For SimCSE, since its official hyper-parameter setting is very different from the other 3 methods, we use the official code\footnote{\url{https://github.com/princeton-nlp/SimCSE}} along with the default hyper-parameters. In detail, its hyper-parameters are: 1 epoch of training, batch size of 512, AdamW optimizer with learning rate 5e-5 and a linear layer on the CLS token embedding as the pooling method.

Since in the real-world scenario where the labeled data is expensive to obtain, applying early-stopping with a in-domain development set is impractical. Thus, in our unsupervised experiments, we do not use early-stopping with in-domain labeled data and indicate a fixed number of training steps\footnote{For SimCSE, the official code involves early-stopping on the STS-B development set. We do not change this setting for this method, since STS-B is not an in-domain dataset in our task- and domain-specific evaluation.} mentioned above instead.

We use the repository of sentence-transformers\footnote{\url{https://github.com/UKPLab/sentence-transformers}} (version number: v0.3.8) to train the in-domain supervised models. For them, the number of training epochs is 10; the maximum number of training steps is 20K; the batch size is 64; the similarity function $\sigma$ is set to cosine similarity; early-stopping is applied by checking the validation performance. To eliminate the influence of randomness, we report the scores averaged over 5 random seeds for all the in-domain unsupervised and supervised models. All the pre-trained checkpoints used are listed in Table~\ref{tbl:checkpoints}.

For BM25, we use the implementation available on Elasticsearch\footnote{\url{https://www.elastic.co/}} with the default settings.

\begin{table}[H]
\centering
\resizebox{14.5cm}{!}{
\begin{tabular}{|l|l|} 
\hline
\textbf{Model Name}         & \textbf{URL}                                                                                    \\ 
\hline
DeCLUTR-base      & https://huggingface.co/johngiorgi/declutr-base 
   \\
ELECTRA-base       & https://huggingface.co/google/electra-base-discriminator                      \\
DistilRoBERTa-base & https://huggingface.co/distilroberta-base                                    \\
RoBERTa-base       & https://huggingface.co/roberta-base                                           \\
DistilBERT-base    & https://huggingface.co/distilbert-base-uncased                                 \\
BERT-base          & https://huggingface.co/bert-base-uncased                                       \\
SBERT-base-nli-v2 & https://huggingface.co/kwang2049/SBERT-base-nli-v2 \\
SBERT-base-nli-stsb-v2 & https://huggingface.co/kwang2049/SBERT-base-nli-stsb-v2 \\
SDRoBERTa-para     & https://huggingface.co/sentence-transformers/paraphrase-distilroberta-base-v1  \\
USE-large       & https://tfhub.dev/google/universal-sentence-encoder-multilingual-large/3       \\
BART-base & https://huggingface.co/facebook/bart-base \\
T5-base & https://huggingface.co/t5-base \\
\hline
\end{tabular}}
\caption{Model checkpoints used in this work.}
\label{tbl:checkpoints}
\end{table}

\section{Results of Other Checkpoints}
\label{sec:other_checkpoints}

The results of other checkpoints besides \textit{BERT-base-uncased} are shown in Table~\ref{tbl:other_checkpoints}. For all the methods, better results are achieved by using BERT checkpoints, which also makes TSDAE significantly outperforms others. We suppose this advantage comes from the additional pre-training task, next sentence prediction of the BERT models, which guides the model to learn from sentence-level contexts.

\begin{table}[H]
\centering
\resizebox{13cm}{!}{
\begin{tabular}{|l|c|c|c|c|c|} 
\hline
\textbf{Method}    & \textbf{AskU.}                     & \textbf{CQADup.}                   & \textbf{TwitterP.}                   & \textbf{SciDocs}                       & \textbf{Avg.}                           \\ 
\hline
\multicolumn{6}{|l|}{ \textit{ELECTRA-base} }                                                                                                                              \\ 
\hline
TSDAE     & \uline{56.6 +/- 1.1}          & \uline{8.0 +/- 0.3}           & \uline{69.0 +/- 1.6}          & \uline{66.2 +/- 5.6}          & \uline{49.9 +/- 1.1}           \\
CT        & 50.3 +/- 0.4                  & 5.0 +/- 0.2                   & 66.5 +/- 0.7                  & 46.1 +/- 0.6                  & 41.6 +/- 0.8                   \\
SimCSE        & 50.9 +/- 0.3                  & 6.2 +/- 0.1                   & 61.8 +/- 0.6                  & 49.3 +/- 0.3                  & 42.0 +/- 0.1                   \\
BERT-flow & 51.3 +/- 0.3                  & 5.2 +/- 0.0                   & 62.4 +/- 0.1                  & 41.2 +/- 0.1                  & 38.4 +/- 3.0                   \\ 
\hline
\multicolumn{6}{|l|}{ \textit{DistilRoBERTa-base} }                                                                                                                        \\ 
\hline
TSDAE     & \uline{58.9 +/- 0.5}          & 12.5 +/- 0.1                  & \uline{68.5 +/- 0.5}          & 59.3 +/- 0.6                  & \uline{49.9 +/- 0.4}           \\
CT        & 57.9 +/- 0.8                  & \uline{13.8 +/- 0.3}          & 63.6 +/- 1.4                  & \uline{62.7 +/- 0.5}          & 49.8 +/- 0.4                   \\
SimCSE & 57.1 +/- 0.2                  & 12.2 +/- 0.1                  & 65.6 +/- 0.8                  & 63.4 +/- 0.4                  & 49.6 +/- 0.2                   \\ 
BERT-flow & 56.0 +/- 0.2                  & 11.1 +/- 0.1                  & \uline{68.5 +/- 0.1}                  & 53.0 +/- 0.2                  & 46.9 +/- 0.1                   \\ 
\hline
\multicolumn{6}{|l|}{ \textit{RoBERTa-base} }                                                                                                                              \\ 
\hline
TSDAE     & \uline{58.3 +/- 0.7}          & 12.2 +/- 0.3                  & \uline{70.0 +/- 0.8}          & 61.4 +/- 0.5                  & 50.3 +/- 0.3                   \\
CT        & 56.7 +/- 0.5                  & \uline{14.2 +/- 0.3}          & 69.4 +/- 1.3                  & {63.1 +/- 0.3}          & \uline{50.5 +/- 0.4}           \\
SimCSE & 56.6 +/- 0.5                  & 12.4 +/- 0.2                  & 66.8 +/- 0.7                  & \uline{64.4 +/- 0.3}                  & 50.1 +/- 0.2                   \\ 
BERT-flow & 54.5 +/- 0.2                  & 10.5 +/- 0.1                  & 69.0 +/- 0.1                  & 53.5 +/- 0.2                  & 46.6 +/- 0.1                   \\ 
\hline
\multicolumn{6}{|l|}{ \textit{DistilBERT-base} }                                                                                                                           \\ 
\hline
TSDAE     & \uline{59.2 +/- 0.3}          & \uline{\textbf{14.6 +/- 0.1}} & \uline{\textbf{73.9 +/- 0.3}} & \uline{72.3 +/- 0.9}          & \uline{54.9 +/- 0.2}           \\
CT        & 57.7 +/- 0.8                  & 14.0 +/- 0.3                  & 66.4 +/- 0.4                  & 72.2 +/- 0.7                  & 52.3 +/- 0.3                   \\
SimCSE & 54.8 +/- 0.7                  & 12.3 +/- 0.1                  & 66.8 +/- 0.6                  & 65.9 +/- 0.1                  & 49.9 +/- 0.3                   \\ 
BERT-flow & 55.0 +/- 0.2                  & 11.0 +/- 0.0                  & 65.9 +/- 0.0                  & 70.5 +/- 0.1                  & 50.5 +/- 0.1                   \\ 
\hline
\multicolumn{6}{|l|}{ \textit{BERT-base} }                                                                                                                                 \\ 
\hline
TSDAE     & \uline{\textbf{59.4 +/- 0.3}} & \uline{14.5 +/- 0.1}          & \uline{73.0 +/- 0.4}          & \uline{\textbf{74.0 +/- 0.4}} & \uline{\textbf{55.2 +/- 0.2}}  \\
CT        & 56.3 +/- 0.7                  & 13.3 +/- 0.3                  & 72.5 +/- 0.5                  & 67.6 +/- 0.4                  & 52.4 +/- 0.3                   \\
SimCSE & 55.9 +/- 0.8                  & 12.4 +/- 0.0                   & 68.5 +/- 0.0                  & 65.7 +/- 0.0                  & 50.6 +/- 0.2                   \\
BERT-flow & 53.7 +/- 0.2                  & 9.2 +/- 0.1                   & 69.3 +/- 0.2                  & 64.5 +/- 0.1                  & 49.2 +/- 0.1                   \\
\hline
\end{tabular}}
\caption{Evaluation of different checkpoints using average precision. `+/-' separates the mean value and standard deviation over scores of 5 random seeds. Best results within each group are underlined and the overall best results are bold.}
\label{tbl:other_checkpoints}
\end{table}



\section{Equivalent Labeling Work} 
The goal of unsupervised sentence embedding learning methods is to eliminate the need of labeled training data, which can be expensive in the creation. However, as shown in \autoref{sec:results}, approaches with sufficient in-domain labeled data significantly outperform unsupervised approaches. 

As far as we know, previous work did not study the point of intersection between unsupervised and supervised approaches: If you only need few labeled examples to outperform unsupervised approaches, annotating those might be the more viable solution.

To find this intersection point, we train the in-domain supervised SBERT approach with varying size of labeled training data. Results are shown in \autoref{fig:pretraining_effect_lg}. To estimate the intersection with more precision, we apply binary search. We set the search precision to the standard deviation of the target score over 5 random seeds.

The results are shown in \autoref{tbl:equivalent_labeling_work}. To match the performance of TSDAE, 140 - 6k annotated examples are required. CQADupStack and the TwitterParaphrase corpus, which compromise various domains, require more labeled data than AskUbuntu (1 domain). Surprisingly, SciDocs, which includes data from all type of scientific domains, the in-domain supervised approach outperforms unsupervised approaches with just 464 labeled examples. This dataset appears to be especially challenging for unsupervised approaches, as we observe a large performance gap between in-domain supervised and unsupervised approaches.

In an annotation experiment on the Twitter dataset, we measured that annotating 100 Tweet pairs takes about 20 minutes for an (experienced) annotator. Hence, the state-of-the-art unsupervised TSDAE approach achieves the same performance as a supervised approach with  0.5 - 20 hours of annotation work for one annotator (2.5h - 100h for 5 crowd annotators).  

\begin{table}[H]
\centering
\resizebox{7cm}{!}{
\begin{tabular}{|c|c|c|c|c|} 
\hline
\textbf{AskU.} & \textbf{CQADup.} & \textbf{TwitterP.} & \textbf{SciDocs} & \textbf{Avg.}  \\ 
\hline
140       & 2661        & 6067        & 464     & 2333  \\
\hline
\end{tabular}}
\caption{Intersection point (number of labeled sentence pairs) between unsupervised TSDAE and in-domain supervised SBERT.}
\label{tbl:equivalent_labeling_work}
\end{table}

\section{Usage for Pre-Training}
\label{sec:pre_training_detailed}
\begin{figure}[H]
  \centering
  \subfloat[AskUbuntu]{\includegraphics[width=70mm]{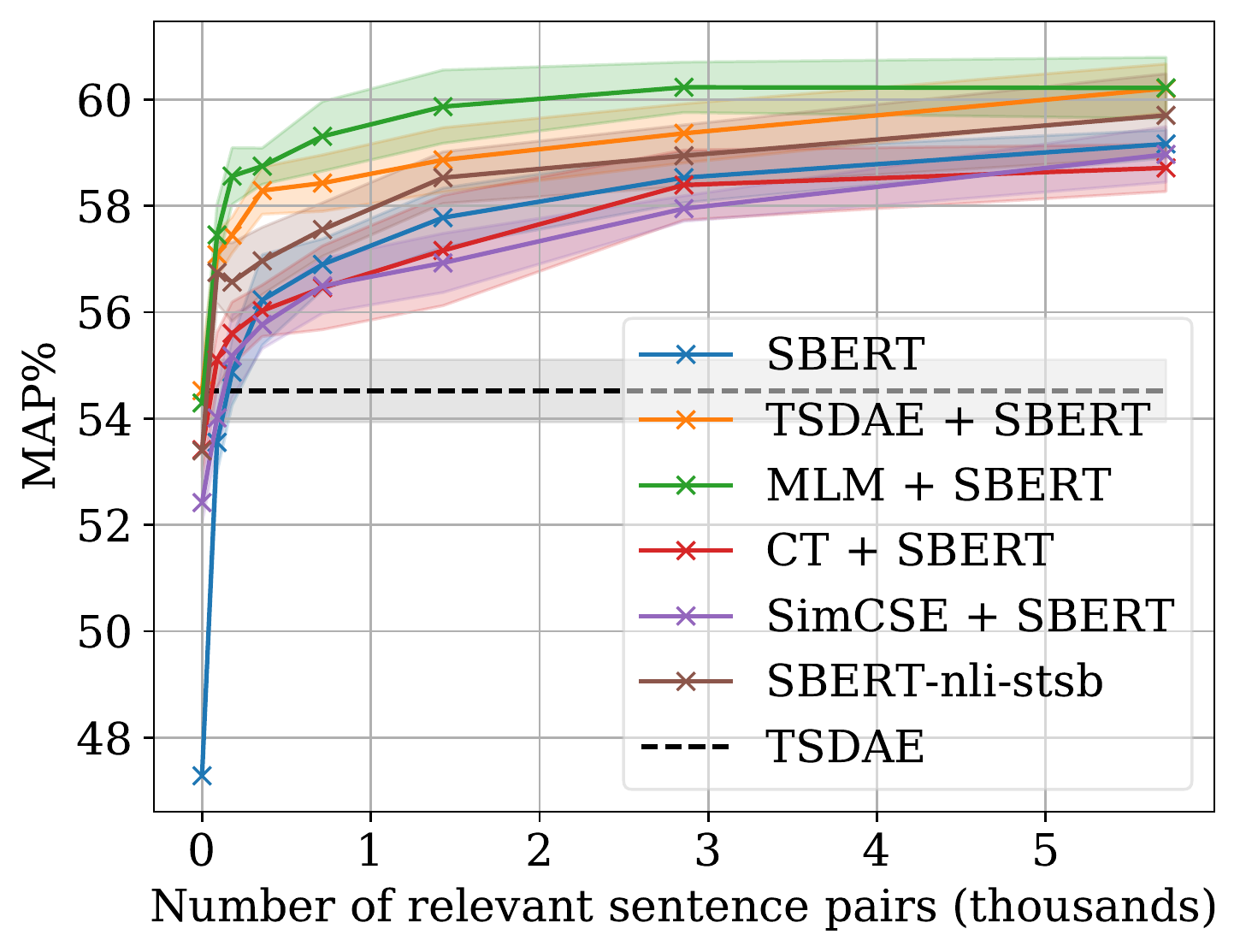}}
  \quad\quad
  \subfloat[CQADupStack]{\includegraphics[width=70mm]{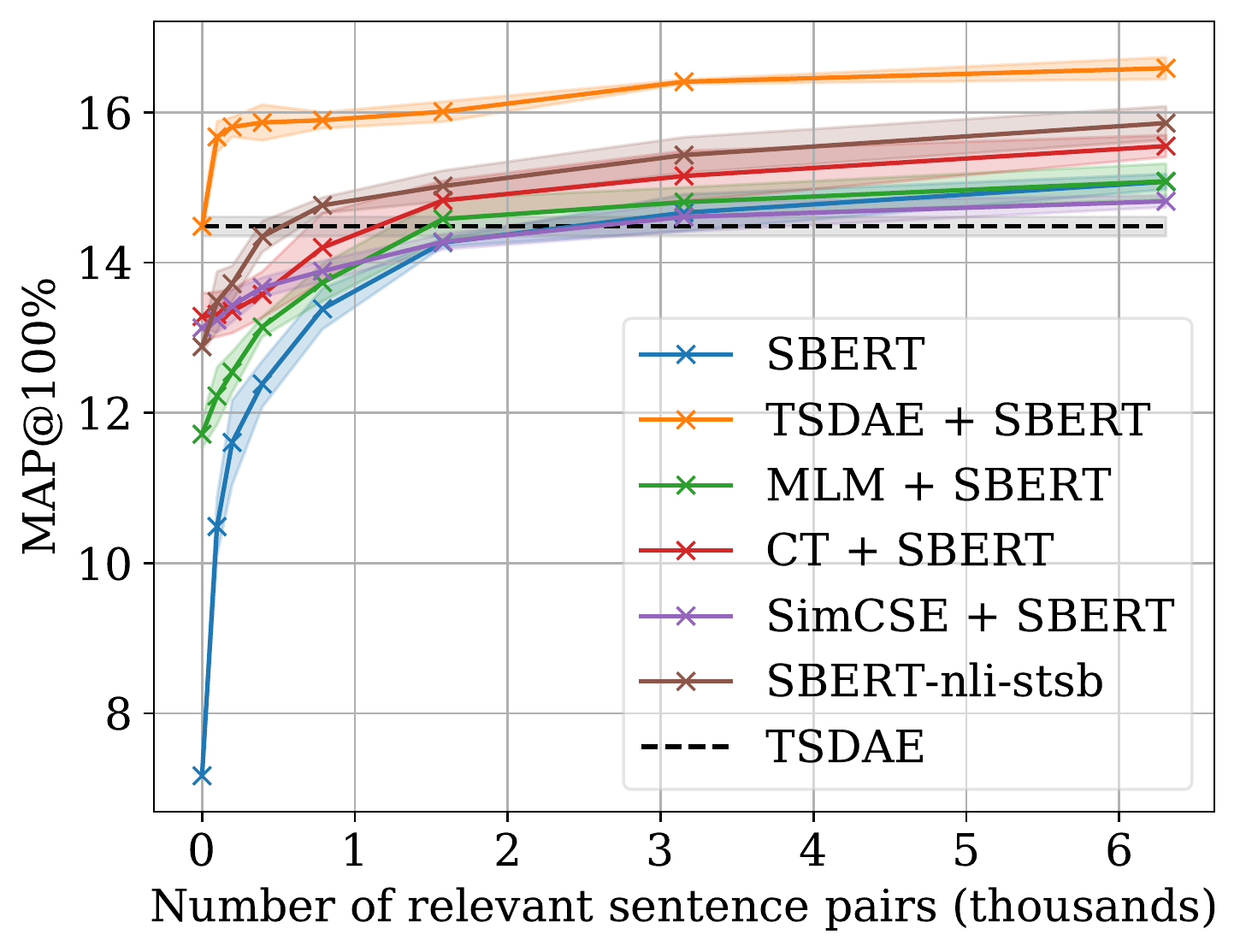}} \\
  \subfloat[TwitterPara]{\includegraphics[width=70mm]{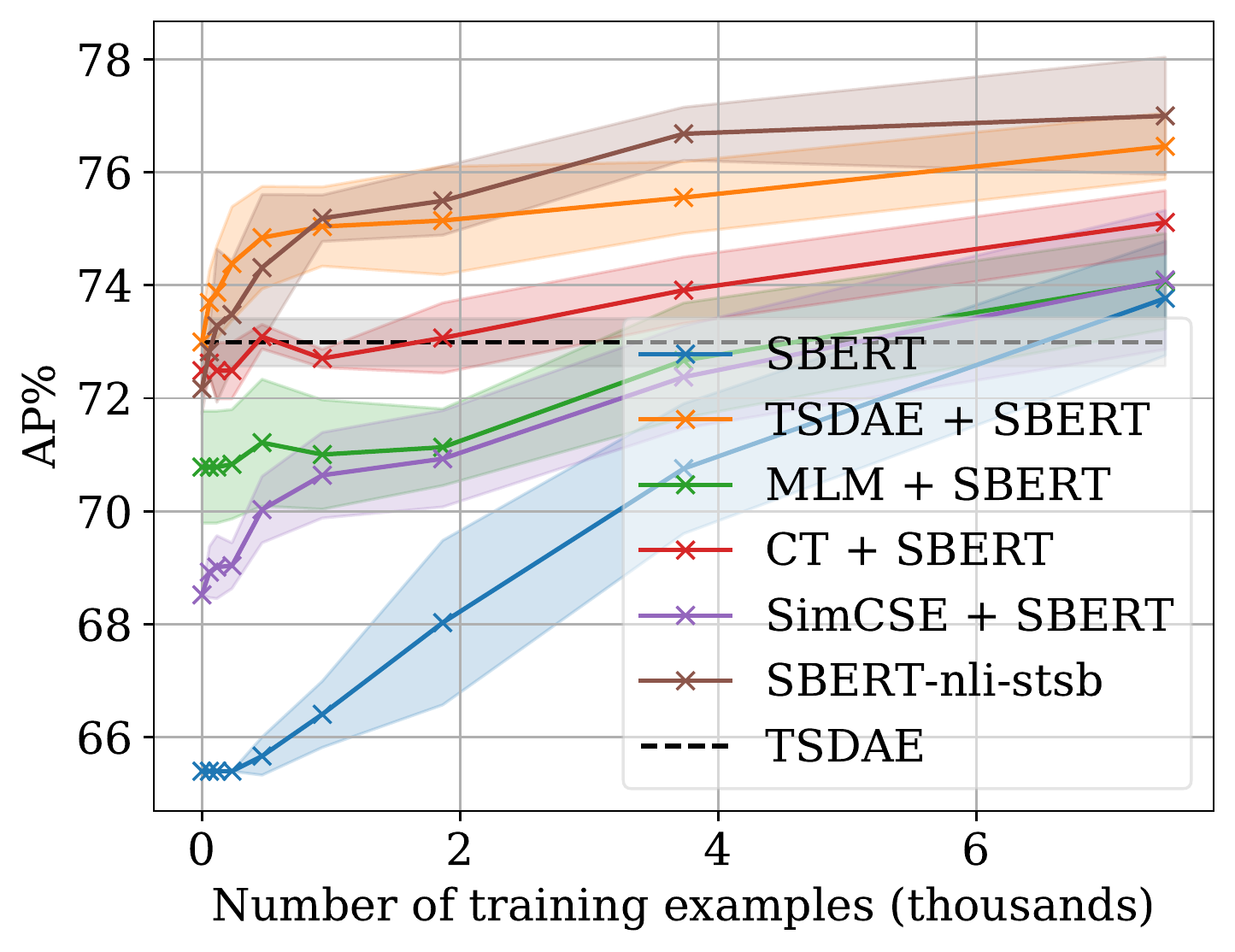}} 
  \quad\quad
  \subfloat[SciDocs]{\includegraphics[width=70mm]{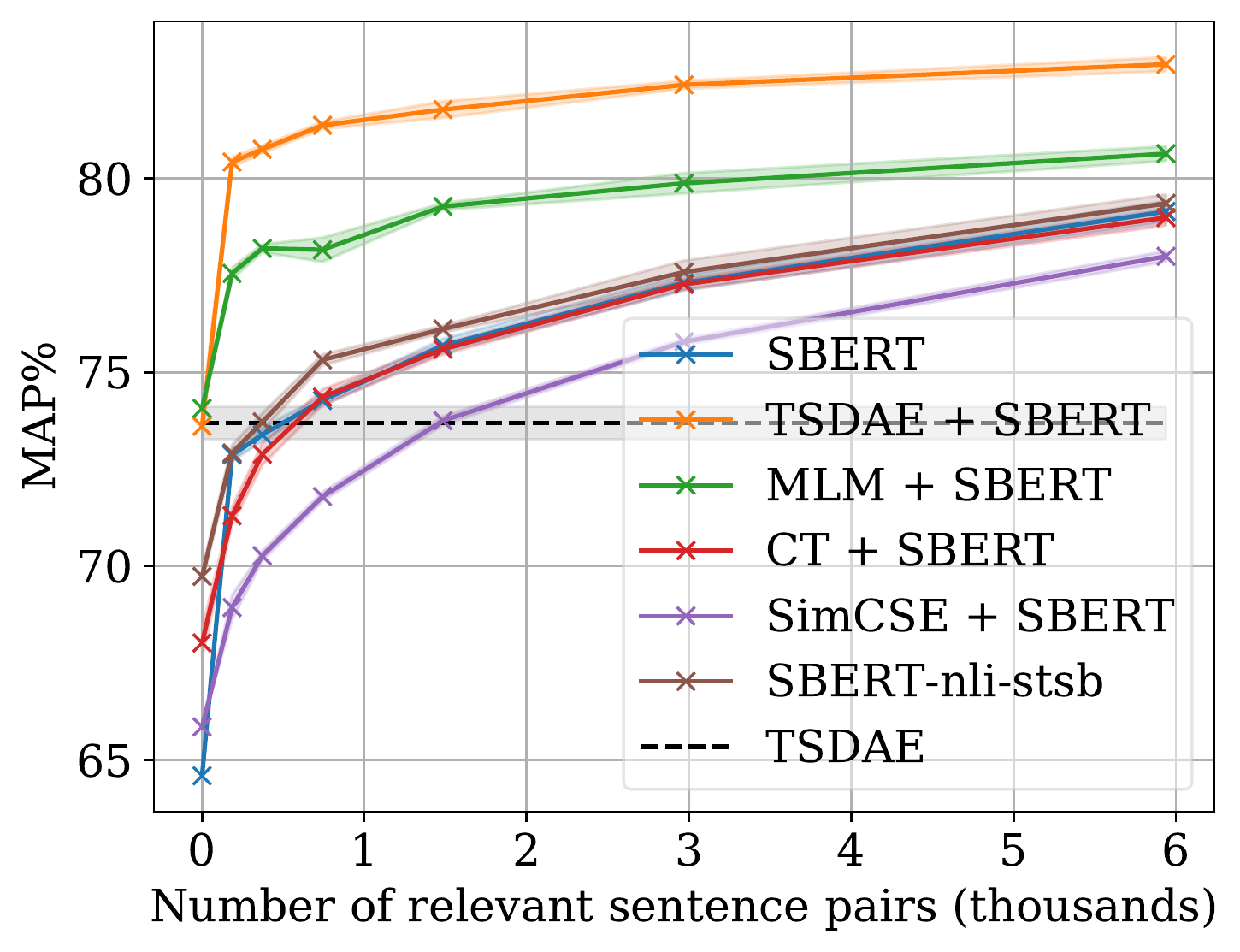}}
  \caption{The influence of the number of training sentences on the model performance.}
  \label{fig:pretraining_effect_lg}
\end{figure}

The pre-training performance on AskUbuntu, CQADupStack and TwitterPara is shown in Figure~\ref{fig:pretraining_effect_lg}.

\section{Detailed Results of Semantic Textual Similarity}
\label{sec:detailed_sts}

The detailed results of STS on each dataset are shown in \autoref{tbl:sts_spearman} with the evaluation metric of Spearman's rank correlation. Note that the training set of STSb contains subsets of STS12-16, Thus, we do not include the scores of SBERT-base-nli-stsb-v2 on these datasets for reducing misunderstanding.

\begin{table}[H]
\centering
\resizebox{13cm}{!}{
\begin{tabular}{|l|c|c|c|c|c|c|c|c|} 
\hline
\textbf{Method}         & \textbf{STS12}                 & \textbf{STS13}                 & \textbf{STS14}                 & \textbf{STS15}                 & \textbf{STS16}                 & \textbf{STSb}                  & \textbf{SICK-R}                & \textbf{Avg.}                   \\ 
\hline
\multicolumn{9}{|l|}{ \textit{Unsupervised method based on BERT-base} }                                                                                                                                                 \\ 
\hline
TSDAE          & 55.2                   & 67.4                   & 62.4                   & 74.3                   & 73.0                   & 66.0                   & 62.3                   & 65.8                    \\
CT             & 60.0           & 76.3           & 68.2           & 77.3           & 75.8                   & 73.9           & 69.4           & 71.6            \\
SimCSE      & 63.6                   & 79.3                   & 69.6                   & 78.2                   & 77.7                   & 73.8                   & 70.1                   & 73.2                    \\
BERT-flow      & 34.1                   & 60.7                   & 48.8                   & 61.9                   & 64.8                   & 48.9                   & 58.4                   & 53.9                    \\
MLM   & 30.9                   & 59.9                   & 47.7                   & 60.3                   & 63.7           & 47.3                   & 58.2                   & 52.6                    \\ 
\hline
\multicolumn{9}{|l|}{ \textit{Out-of-the-box supervised pre-trained models} }                                                                                                                                           \\ 
\hline
SBERT-base-nli-v2 & 72.5 & \textbf{84.8} & \textbf{80.2} & \textbf{84.8} & \textbf{80.0} & 83.9 & 78.0 & \textbf{80.6} \\
SBERT-base-nli-stsb-v2 & -- & -- & -- & -- & -- & \textbf{87.3} & \textbf{80.4} & -- \\
USE-large      & \textbf{{74.3}}  & 71.8                   & 71.4                   & 82.5                   & 77.5                   & 80.9                   & {{75.8}}  & 76.3                    \\
\hline
\end{tabular}}
\caption{Evaluation on the task of STS using Spearman's rank correlation.}
\label{tbl:sts_spearman}
\end{table}

\section{Influence of Corpus Size}
\label{sec:training_size_lg}
\begin{figure}[H]
  \centering
  \subfloat[AskUbuntu]{\includegraphics[width=70mm]{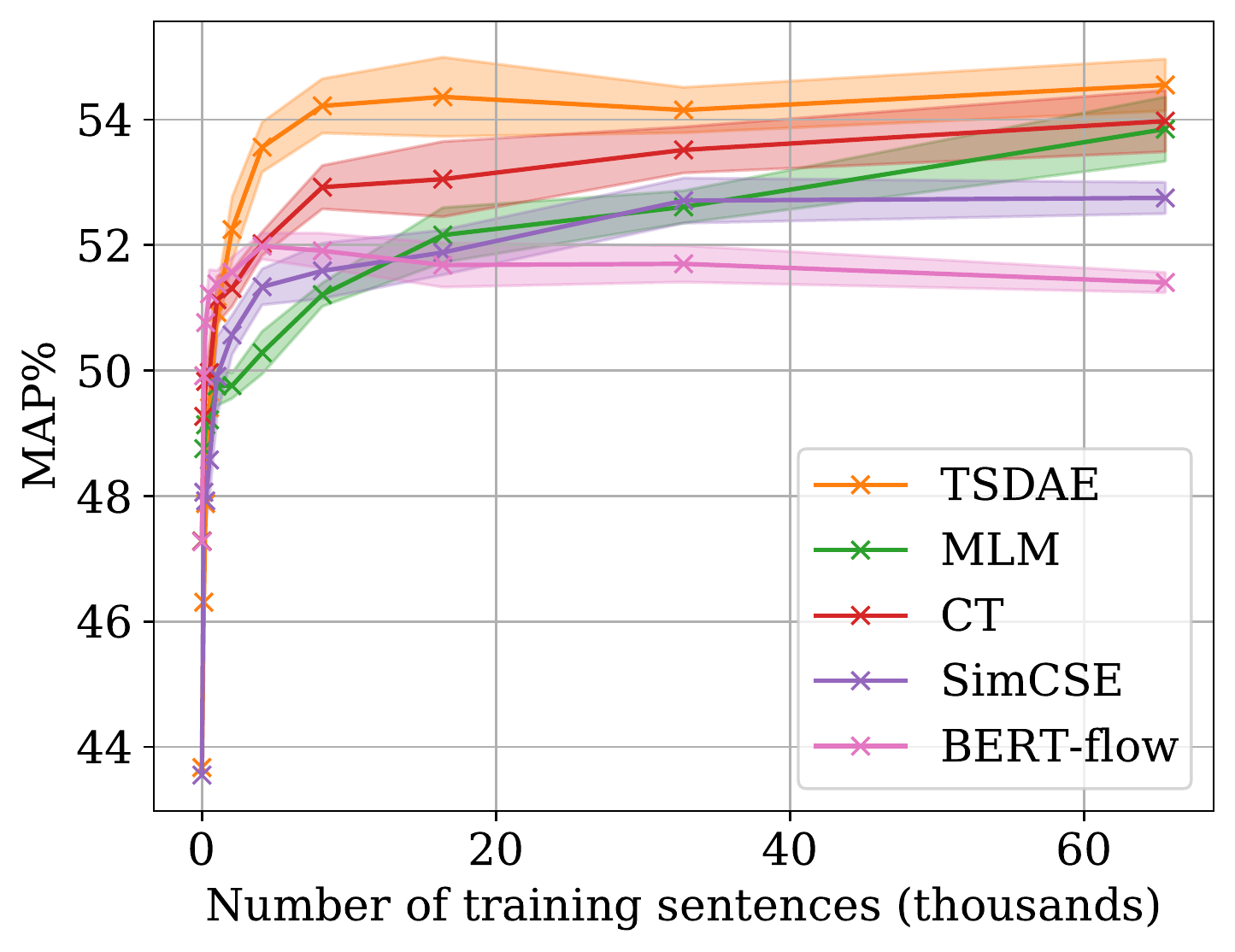}}
  \quad\quad
  \subfloat[CQADupStack]{\includegraphics[width=70mm]{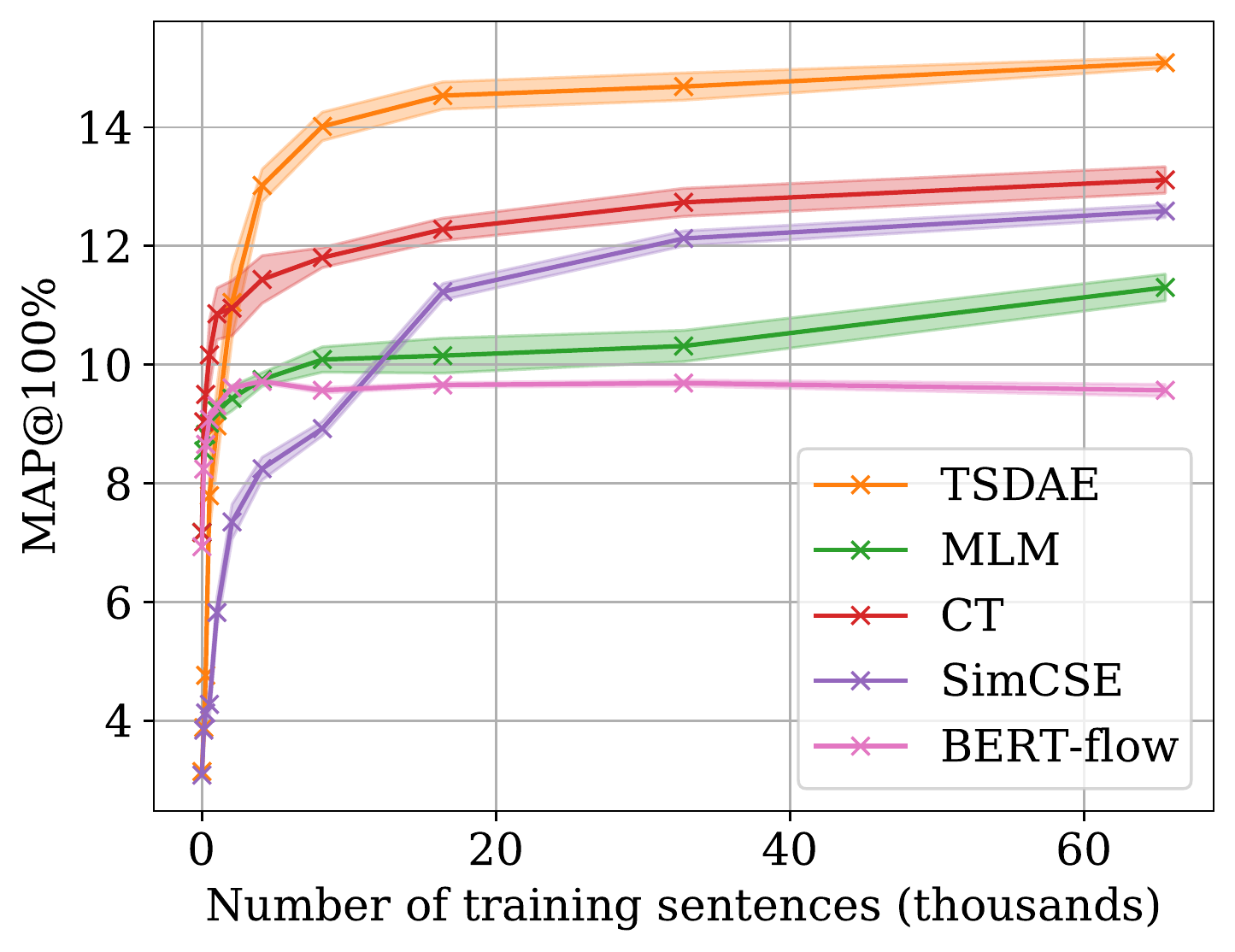}} \\
  \subfloat[TwitterPara]{\includegraphics[width=70mm]{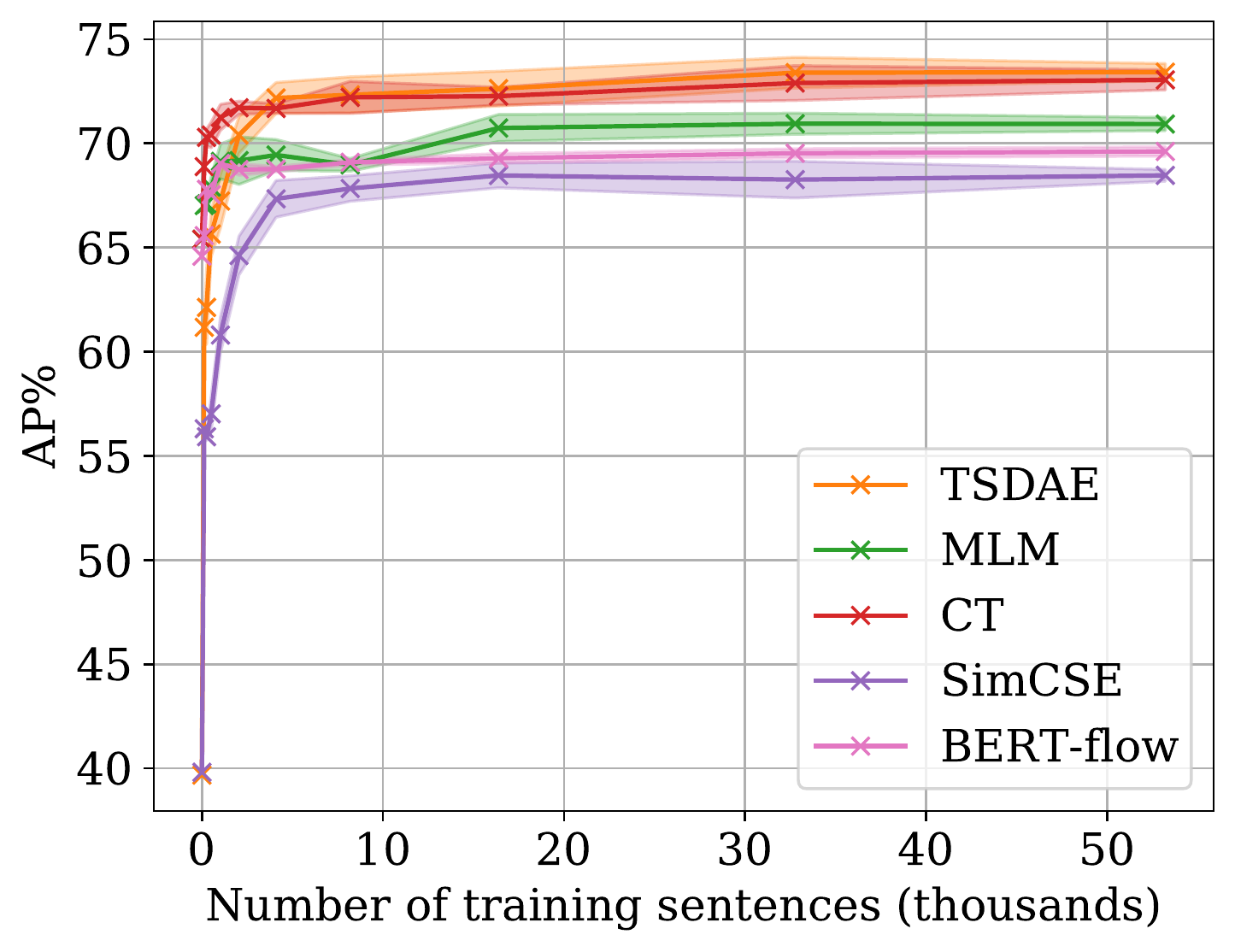}} 
  \quad\quad
  \subfloat[SciDocs]{\includegraphics[width=70mm]{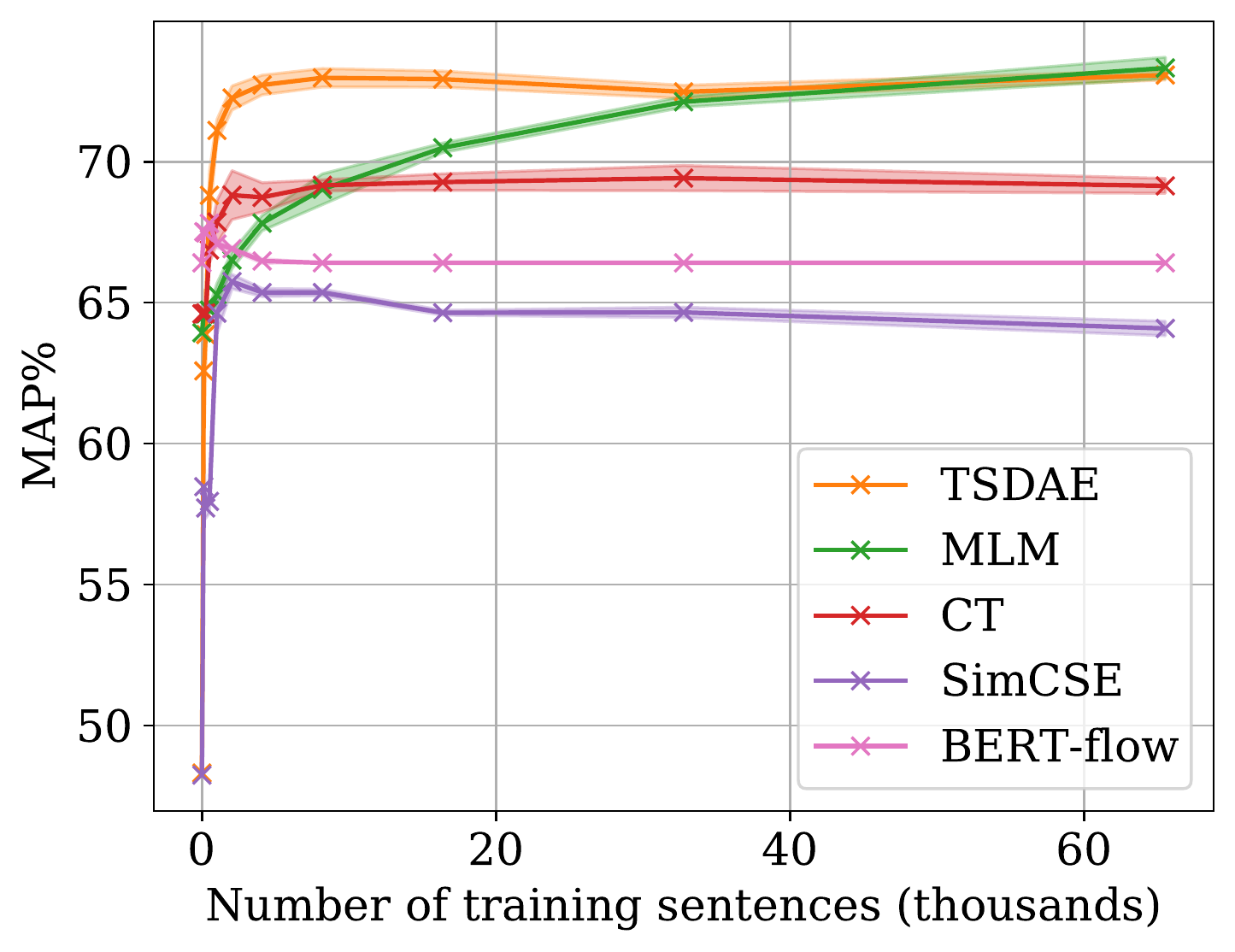}}
  \caption{The influence of the number of training sentences on the model performance.}
  \label{fig:training_size_lg}
\end{figure}

The influence of corpus size for AskUbuntu, CQADupStack and TwitterPara is shown in \autoref{fig:training_size_lg}.

\section{Influence of Different POS Tags}
\label{sec:pos_tags}
\begin{figure}[H]
  \centering
  \subfloat[AskUbuntu]{\includegraphics[width=75mm]{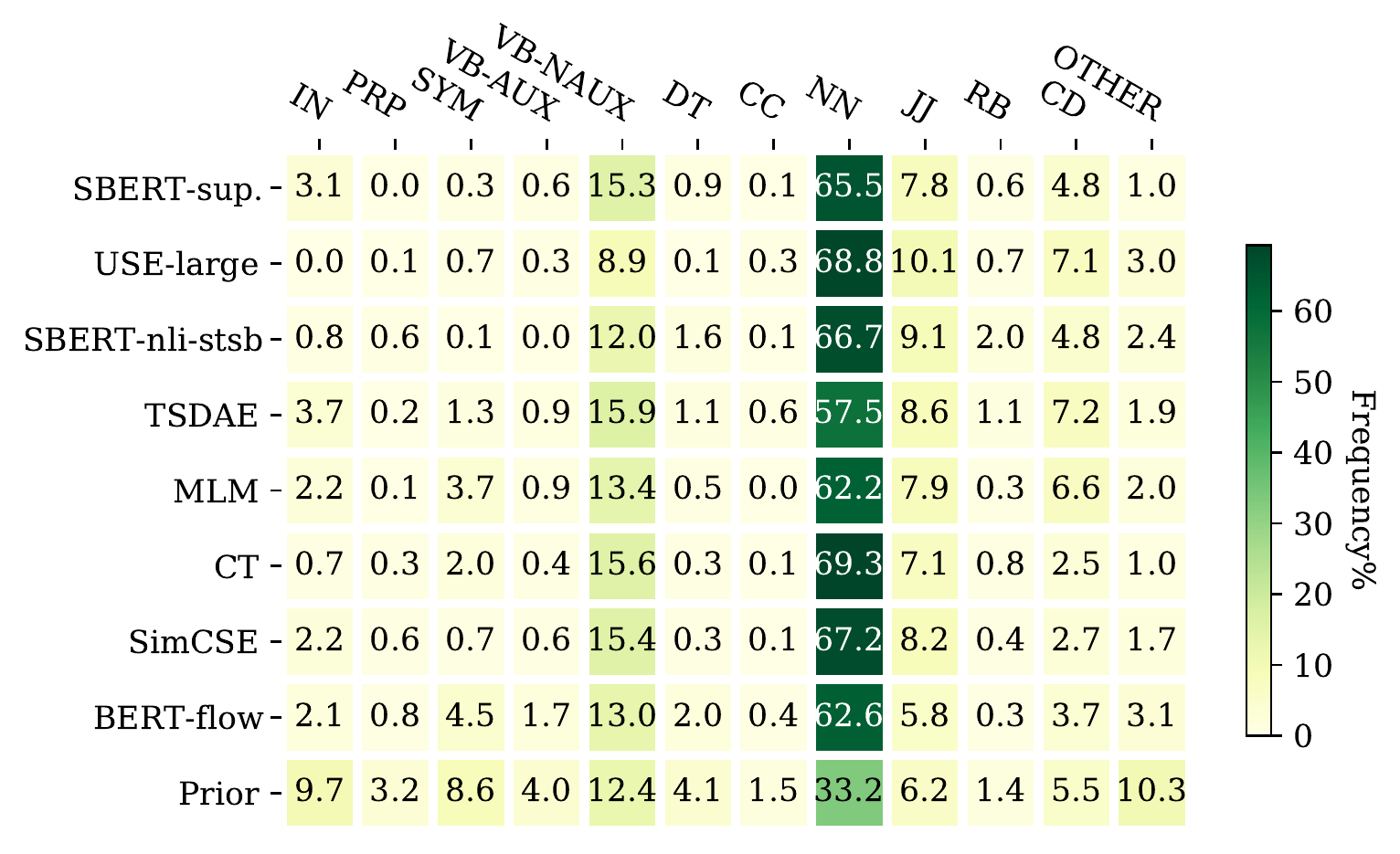}}
  \subfloat[CQADupStack]{\includegraphics[width=75mm]{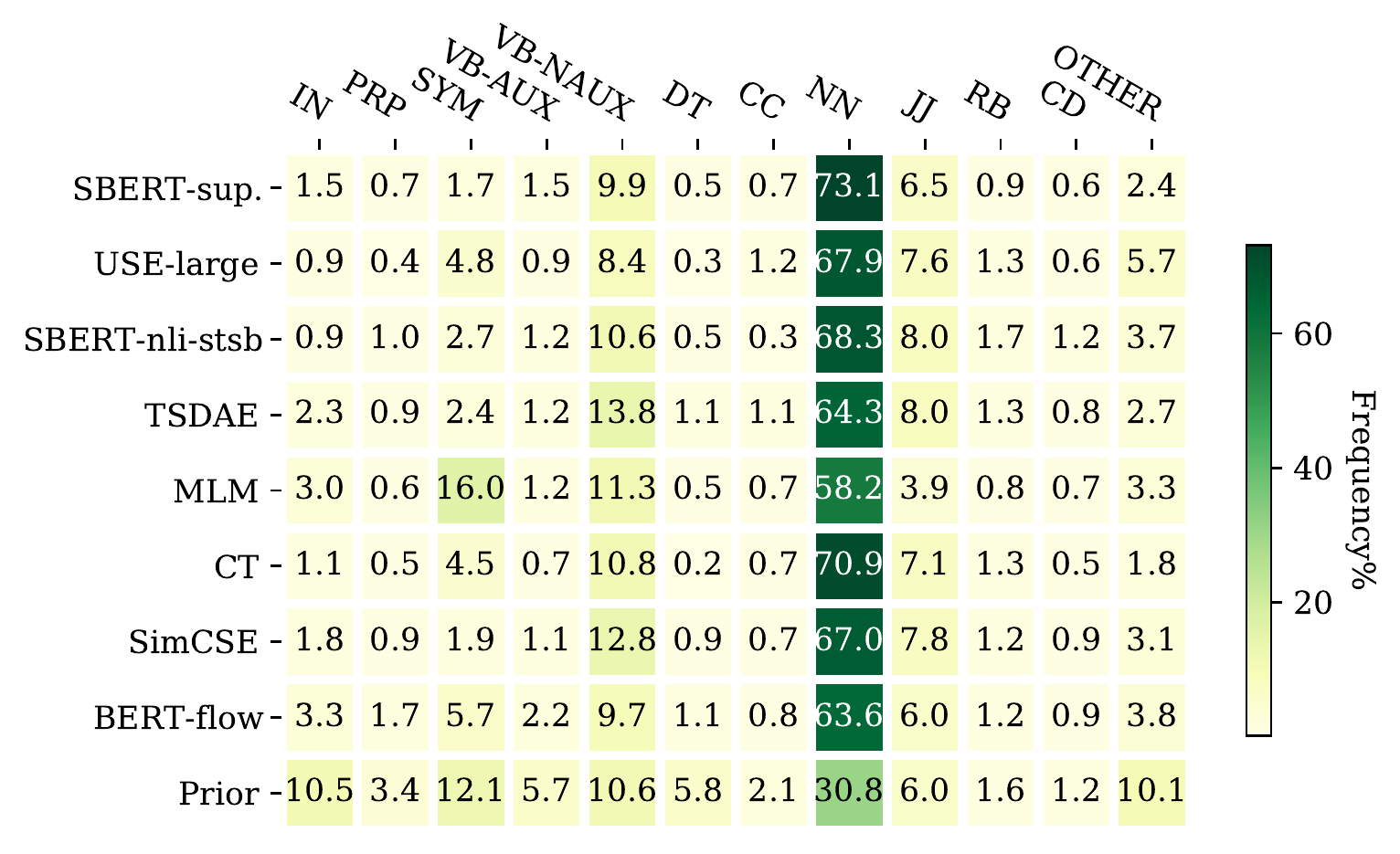}} \\
  \subfloat[TwitterPara]{\includegraphics[width=75mm]{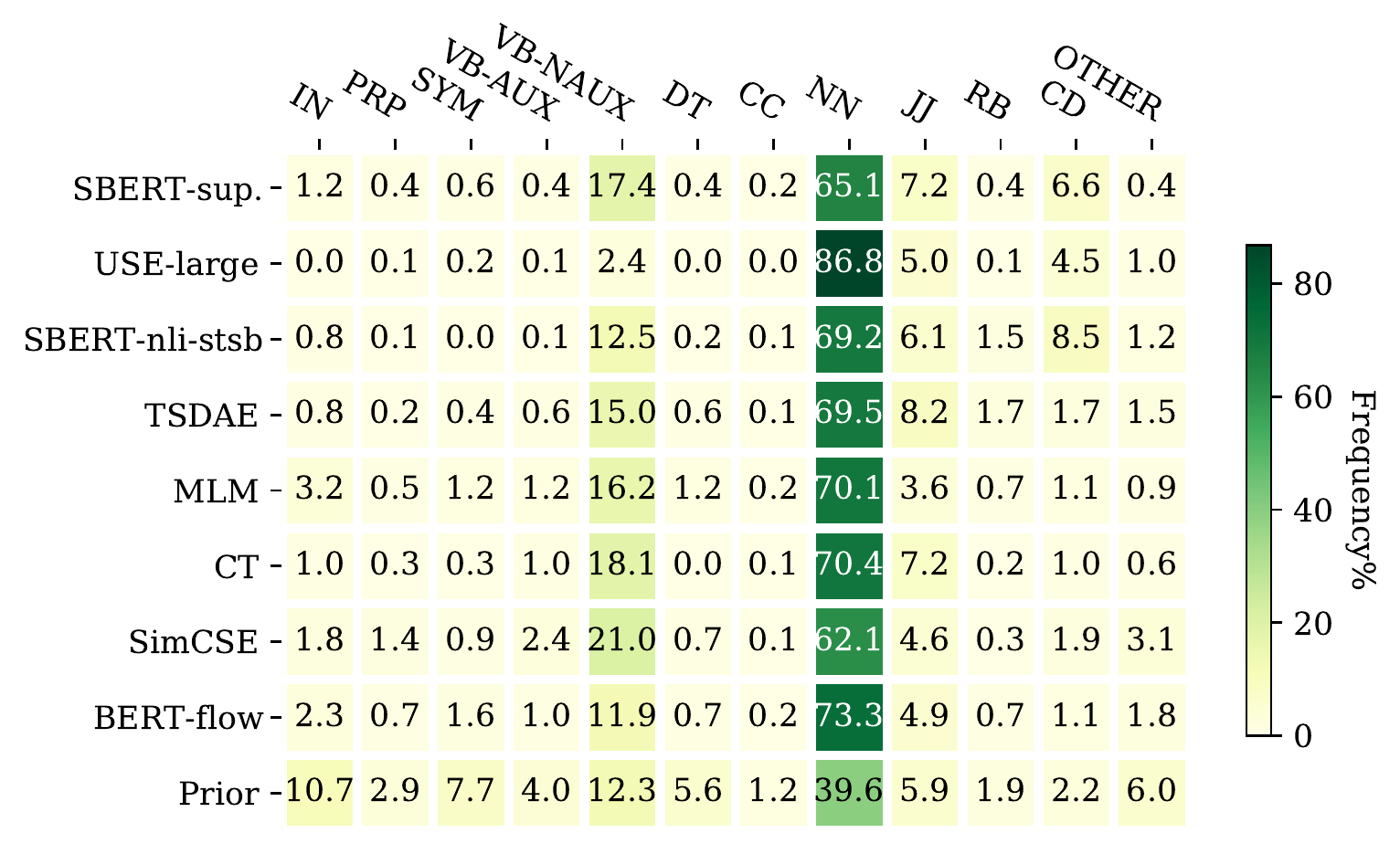}} 
  \subfloat[SciDocs]{\includegraphics[width=75mm]{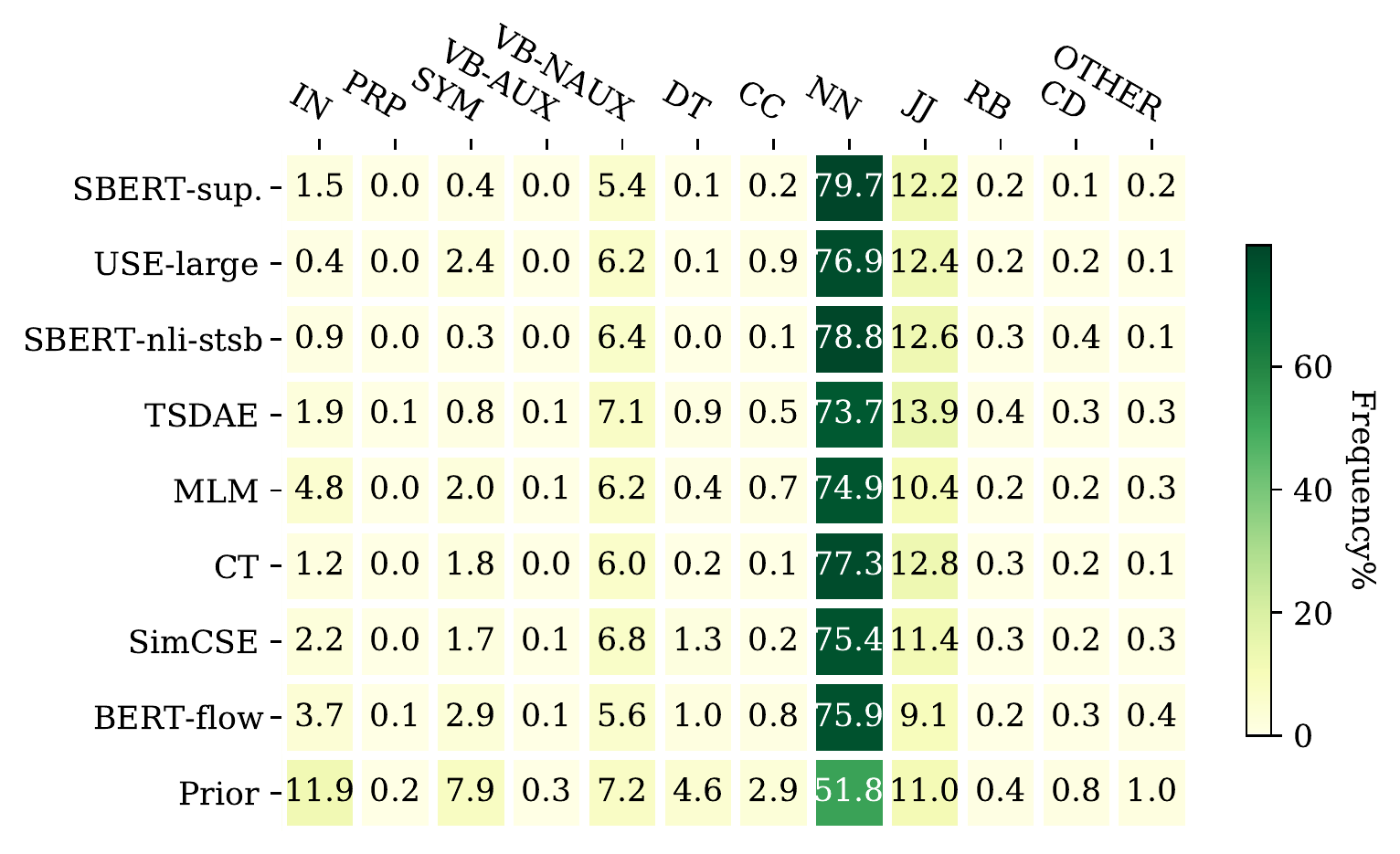}}
  \caption{The influence of different POS tags on the output similarity scores.}
  \label{fig:pos_tags_influence}
\end{figure}

The influence of different POS tags on the output similarity scores for AskUbuntu, CQADupStack and TwitterPara is shown in Figure~\ref{fig:pos_tags_influence}.

\end{document}